\pdfoutput=1
\documentclass[
aip,
amsmath,amssymb,
reprint,
]{revtex4-1}

\usepackage{dcolumn}
\usepackage{bm}
\usepackage[utf8]{inputenc}
\usepackage[T1]{fontenc}
\usepackage{mathptmx}
\usepackage{etoolbox}
\usepackage{url}
\usepackage{hyperref}
\usepackage[sfdefault]{biolinum} 
\usepackage{caption}
\usepackage{multirow} 
\usepackage{graphicx}
\usepackage{booktabs}
\usepackage{cancel}
\usepackage{microtype}

\DeclareCaptionLabelSeparator{pipe}{\space|\space}
\renewcommand{\tablename}{Table}
\renewcommand{\thetable}{\arabic{table}}

\usepackage{hyperref}
\hypersetup{
	colorlinks=true,
	linkcolor=blue,
	citecolor=blue,
	urlcolor=blue
}

\makeatletter
\def\@email#1#2{
	\endgroup
	\patchcmd{\titleblock@produce}
	{\frontmatter@RRAPformat}
	{\frontmatter@RRAPformat{\produce@RRAP{*#1\href{mailto:#2}{#2}}}\frontmatter@RRAPformat}
	{}{}
}
\makeatother

\begin{document}
	
	\preprint{AIP/123-QED}
	
	\title[\small\textsc{Neuronal Constraints Drive Superior Learning in RNNs}]{Harnessing cortical geometry, wiring, and function as inductive biases for recurrent neural networks}
	\author{Mo Shakiba}
	\affiliation{Neuromatch Academy, Neuromatch, Inc., USA}
	
	\author{Rana Rokni}
	\affiliation{Neuromatch Academy, Neuromatch, Inc., USA}
	
	\author{Mohammad Mohammadi}
	\affiliation{Neuromatch Academy, Neuromatch, Inc., USA}
	
	\author{Nima Dehghani}
	\email{nima.dehghani@mit.edu}
	
	\affiliation{McGovern Institute for Brain Research, Massachusetts Institute of Technology (MIT)}

	\date{\today}
	
	
	\begin{abstract}
		How the wiring and functional organization of cortex shape recurrent computation remains a central question in both neuroscience and machine learning. Here, we leverage data released through the Machine Intelligence from Cortical Networks (MICrONS) program—a functional connectomics resource spanning multiple areas of mouse visual cortex, in which dense calcium imaging is co-registered with high-resolution electron microscopy reconstruction from the same animal—to build biologically grounded recurrent neural networks. Using neuronal spatial coordinates, anatomical connectivity, and function-derived relationships from nearly 12,000 coregistered excitatory neurons, we initialize recurrent weights and impose communication-aware spatial constraints during learning. Across three cognitive decision-making tasks, networks constrained by cortical structure and function consistently outperform baseline and partially constrained models. Functional weight initialization provides the largest gain, while real spatial embedding yields robust additional improvements across conditions. These biologically grounded networks also develop low-entropy, modular, and small-world organization, and retain strong performance even when recurrence is restricted to positive weights. Together, our results show that the machinery of cortex—its geometry, wiring, and functional structure—can be harnessed as a powerful inductive basis for building recurrent networks that learn more effectively while converging toward key organizational principles of biological computation.  \footnote{Context $\&$ overview:\\ \url{https://neurovium.science/posts/pblog-Cortical-blueprint-RNN}
			\\ Code $\&$ experiments: \\ \url{https://github.com/neurovium/CorticalBlueprintRNN}}
	\end{abstract}
	
	\maketitle
	
	\section{\label{sec:introduction} Introduction}
	Understanding how cortical organization shapes computation---and how that organization can be used to build better artificial systems---remains a central challenge at the interface of neuroscience and machine learning. Artificial neural networks increasingly serve not only as engineering tools but also as normative models for how biological circuits might implement efficient computation and learning \cite{dayan_comp_2005,richards_deep_2019,marblestone2016towardanintegration}. Recurrent neural networks (RNNs) are especially useful in this context because they support temporal integration, memory, and decision-making, while also providing a flexible framework for testing how circuit organization shapes computation \cite{barak2017recurrentneuralnetworks}. Yet most RNNs are still built from generic initializations and relatively unconstrained connectivity, despite the fact that cortical circuits are not arbitrary: neurons occupy physical space, are shaped by wiring economy, and participate in structured anatomical and functional relationships across scales \cite{bullmore2012theeconomyof,samu2014influenceofwiring,budd2012communicationandwiring,schroter2017microconnectomicsprobingthe,shipp2007structureandfunction,wang2016brainstructureand,Varshney2011celegansconnectome,Chen2017primateconnectome,Reimann2026nonrandomArchitect,Kaiser2007clustercortical}.
	
	A growing body of work has shown that imposing biological constraints on artificial networks can improve both learning and organization \cite{mcallister2026nonrandombrainconnectome,fruengel_sparsity_2025,Liao2024selfassemblybiologicallyplausiblelearning}. Studies at the interface of neuroscience and machine learning have argued that sparse local connectivity, topographic structure, recurrence, inhibition, and other circuit-level constraints should be treated as computationally meaningful design principles rather than merely biological detail \cite{pulvermuller2021biologicalconstraintson,marblestone2016towardanintegration, miconi_hebb_2017, fruengel_sparsity_2025}. In particular, spatially embedded recurrent neural networks (seRNNs) demonstrated that assigning recurrent units positions in Euclidean space and penalizing long-range communication yields sparse, modular, and small-world architectures while preserving strong task performance \cite{achterberg2023spatiallyembeddedrecurrent}. Related spatially constrained sparse RNNs likewise learn faster and more data-efficiently across cognitive tasks than fully connected baselines \cite{khona2023winningthelottery}. These findings suggest that geometry and wiring constraints are not merely biological ornamentation, but useful inductive biases for recurrent computation.
	
	However, most biologically inspired RNNs still rely on stylized or randomized embeddings rather than the measured organization of real cortical circuits. Standard RNN practice remains dominated by random initialization and unconstrained recurrent wiring, with structure emerging only through training \cite{schuessler2020theinterplaybetween,krause2022operativedimensionsin}. This leaves an important question unresolved: \emph{do recurrent networks benefit merely from generic biological inspiration, or can we harness the specific geometry, wiring, and functional structure of cortex as strong inductive biases that can drive superior learning?}
	
	This question has become newly tractable with the emergence of multimodal functional connectomics. Data released through the Machine Intelligence from Cortical Networks (MICrONS) program pair dense calcium imaging with high-resolution electron microscopy reconstruction in the same mouse visual cortex, linking spatial position, anatomical connectivity, and functional activity within a common cortical substrate \cite{turner2020multiscaleandmultimodal,bae2025functionalconnectomicsspanning}. In the subset used here, this framework provides nearly 12,000 coregistered excitatory neurons, making it possible to move beyond abstract spatial embedding and instead ground recurrent models directly in measured cortical geometry, wiring, and function. More broadly, resources such as MICrONS are valuable not only because they describe cortical circuits, but because they make those circuits usable as architectural substrates for machine learning. Co-registered structure–function datasets allow recurrent models to be constrained by measured geometry, connectivity, and activity within the same neuronal population, rather than by abstract priors alone \cite{johnson2023exploitinglargeneuroimagingdatasets}.
	
	This problem is especially important because recent connectome-constrained modeling studies suggest that anatomy alone may not uniquely specify recurrent dynamics \cite{Seung2024predictfunctionfromstructure}. Empirical connectivity can provide a powerful scaffold for predicting neural activity, yet multiple dynamical solutions may remain compatible with the same wiring unless additional physiological or functional constraints are included \cite{lappalainen2024connectomeconstrainednetworkspredict,wang2016brainstructureand,beiran2025predictionofneural}. A more informative strategy, therefore, is to combine anatomical structure with functional measurements from the same neuronal population, constraining recurrent models not only by where neurons are and how they are wired, but also by how they co-vary during activity.
	
	Here, we use MICrONS-derived cortical geometry, wiring, and function to construct a family of biologically grounded recurrent neural networks. Functional relationships derived from neuronal activity inform recurrent weight initialization, while measured spatial coordinates and connectivity-derived communication structure shape spatial regularization during learning. We compare eleven model variants that selectively include or omit these biological priors and train them on three cognitive decision-making tasks. This design allows us to disentangle the contributions of biologically informed initialization, real spatial embedding, and communicability-based regularization to both task performance and emergent graph-theoretic organization.
	
	We show that cortical structure--function priors provide a powerful inductive basis for recurrent computation. Across tasks, biologically grounded models outperform baseline and partially constrained variants. Functional initialization yields the largest performance gains, whereas real spatial embedding confers reliable additional improvements and steers learned networks toward lower-entropy, more modular, and more small-world connectivity. These advantages remain evident even when recurrence is restricted to positive weights. Together, our results show that the machinery of cortex can be used not only to build more powerful recurrent networks, but also to identify which features of cortical organization are most useful for learning.

	\begin{figure*}
		\includegraphics[width=\textwidth]{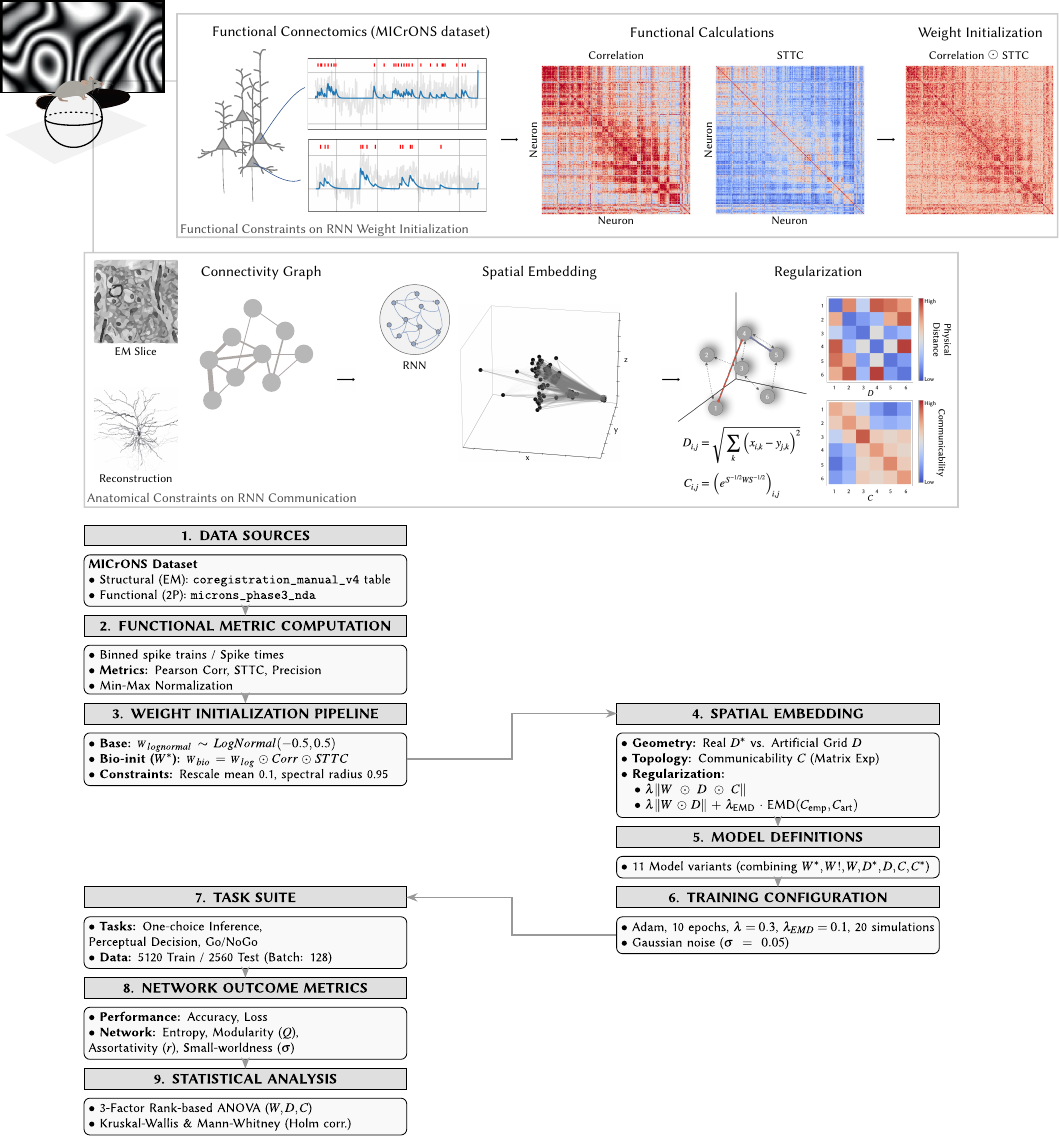}
		\caption{\label{fig:overview}\textbf{Conceptual framework illustrating how the MICrONS dataset was used to constrain RNN models.} Functional data from two-photon calcium imaging were used to compute pairwise correlation and STTC matrices for biologically grounded weight initialization. EM-derived anatomical data and neuronal coordinates were used to construct the spatially embedded graph from which distance and communicability terms were derived for regularization.}
	\end{figure*}
	
	\section{\label{sec:results} Results}
	
	\subsection{\label{sec:taskperf} Cortical structure--function priors improve recurrent learning across tasks}
	
	We first asked whether recurrent networks grounded in measured cortical geometry, wiring, and function learn more effectively than unconstrained or partially constrained alternatives. To do so, we compared eleven RNN variants that selectively incorporated biologically informed weight initialization (\textit{W*} or \textit{W!}), real neuronal coordinates from MICrONS (\textit{D*}), and communicability-based regularization (\textit{C} or \textit{C*}) (Table~\ref{tab:model_table}). Each model was trained across 20 runs on three cognitive decision-making tasks. This design allowed us to treat cortical inductive biases not as a single intervention, but as separable components whose effects on learning could be systematically evaluated.
	
	Across all three tasks, models constrained by cortical structure and function consistently outperformed baseline and partially constrained models (Table~\ref{tab:acc_table}). The strongest overall performance was achieved by the fully biologically grounded variants, \textit{W*D*C} and \textit{W*D*C*}, which combine function-derived weight initialization with real spatial embedding and communicability-aware regularization. The \textit{W*D*C} model reached mean accuracies of 0.917, 0.865, and 0.948 on Tasks 1--3, respectively, while \textit{W*D*C*} achieved 0.985, 0.880, and 0.951. By contrast, minimally constrained or partially constrained models performed substantially worse, particularly on Tasks 2 and 3, where several variants remained near chance. These results show that cortical priors do not merely reshape network organization; they materially improve recurrent learning.
	
	The structure of the ablation results clarifies the contribution of each cortical prior. When spatial embedding was fixed to real neuronal coordinates (\textit{D*}), performance differed strongly across weight-initialization conditions, with both biologically initialized variants (\textit{W*} and \textit{W!}) significantly outperforming standard initialization across all tasks. Crucially, no significant difference was observed between \textit{W*} and \textit{W!}, indicating that the advantage does not depend on preserving an exact one-to-one mapping between neuronal position and initial weight assignment. Instead, the gain appears to arise primarily from the biologically derived weight statistics themselves. Thus, function-derived initialization provides the strongest and most general inductive bias for learning in these recurrent networks.
	
	Real spatial embedding contributed an additional, distinct benefit. When weight initialization and communicability formulation were held fixed, replacing artificial grid coordinates (\textit{D}) with real neuronal coordinates (\textit{D*}) significantly improved performance across tasks. This effect was especially clear in models lacking biological initialization, showing that cortical geometry itself provides useful structure for recurrent optimization. The actual spatial arrangement of cortical neurons therefore acts as a meaningful computational prior rather than a purely anatomical detail.
	
	The effect of communicability was subtler and more context dependent. Different communicability formulations produced smaller and more task-specific changes than either weight initialization or spatial embedding. In Task 1, these differences were limited. In Tasks 2 and 3, however, direct communicability regularization (\textit{C}) generally improved performance relative to either the EMD-based formulation (\textit{C*}) or the unregularized condition, particularly when combined with real spatial embedding. Communicability is therefore not the primary driver of performance, but instead refines learning once the model is already grounded in biologically meaningful initialization and geometry. The exact Kruskal–Wallis statistics and Holm‑corrected post hoc pairwise comparisons underlying these ablation results are reported in Supplementary Results  (\emph{Effect of \textit{W} for fixed \textit{D}, \textit{C}}, \emph{Effect of \textit{D} for fixed \textit{W}, \textit{C}}, and \emph{Effect of \textit{C} for fixed \textit{W}, \textit{D}}) and in Supplementary Tables~\ref{tab:S5_mw_pairwise_W}--\ref{tab:S7_mw_pairwise_C}).
	
	Taken together, these comparisons reveal a clear hierarchy among cortical priors. Functional initialization provides the largest performance gain, real spatial embedding adds a robust secondary benefit, and communicability contributes more selectively depending on task and architectural context. This hierarchy is important scientifically because it helps isolate which aspects of cortical organization are most useful for recurrent computation, and it is important for designing better artificial systems because it shows that biologically informed inductive biases can improve learning without increasing architectural complexity.
	
	\begin{figure*}
		\includegraphics[width=\textwidth]{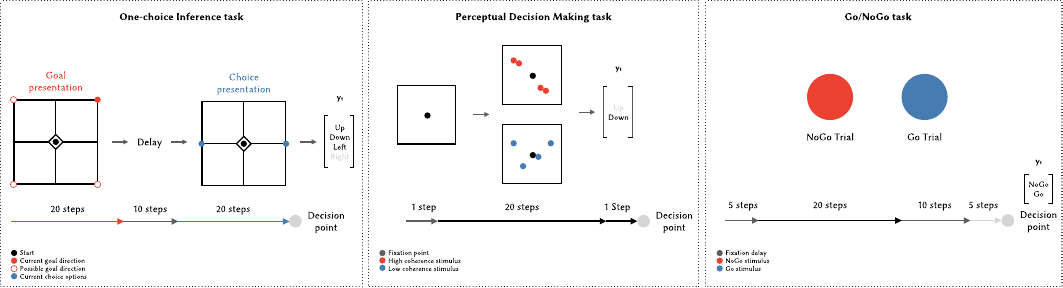}
		\caption{\label{fig:tasks}\textbf{Task paradigms used to train RNN models.} Left: One-Choice Inference task, where the network integrates goal (red) and choice (blue) stimuli across a delay to select the correct movement direction. Middle: Perceptual Decision-Making task, where the network identifies the dominant direction in noisy stimuli (high coherence: red, low coherence: blue). Right: Go/NoGo task, where the network must either respond (blue) or withhold a response (red) following stimulus presentation and a short delay. Arrows indicate the temporal sequence of stimulus presentation and decision points.}
	\end{figure*}
	
	\subsection{\label{sec:robustness} Functional priors stabilize learning under positive-only recurrence}
	
	We next asked whether biologically grounded initialization also improves robustness under a more restrictive recurrent architecture. To test this, we constrained recurrent weights to be strictly positive, mimicking an excitatory-only regime, and evaluated performance across model variants (Table~\ref{tab:acc_table_sign}). Under this constraint, all randomly initialized models collapsed to near-chance performance across tasks. In striking contrast, models initialized with functional priors (\textit{W*} and \textit{W!}) retained high accuracy, often remaining near ceiling.
	
	These results show that functional initialization does more than improve average performance under standard training conditions. It also stabilizes learning when the recurrent architecture is made substantially more difficult to optimize. This is consistent with the idea that biologically derived initial weight structure biases optimization toward more favorable regions of parameter space, thereby preserving trainability even when recurrent dynamics are sign-constrained.
	
	Importantly, the permuted-weight variant \textit{W!} performed comparably to \textit{W*} under this positive-only regime, reinforcing the conclusion that robustness depends primarily on the statistical structure of the biologically derived weight distribution rather than on preserving a precise neuron-to-weight mapping.
	
	\subsection{\label{sec:topology} Cortical priors drive emergent network organization toward brain-like topology}
	
\begin{figure*}
		\includegraphics[width=\textwidth]{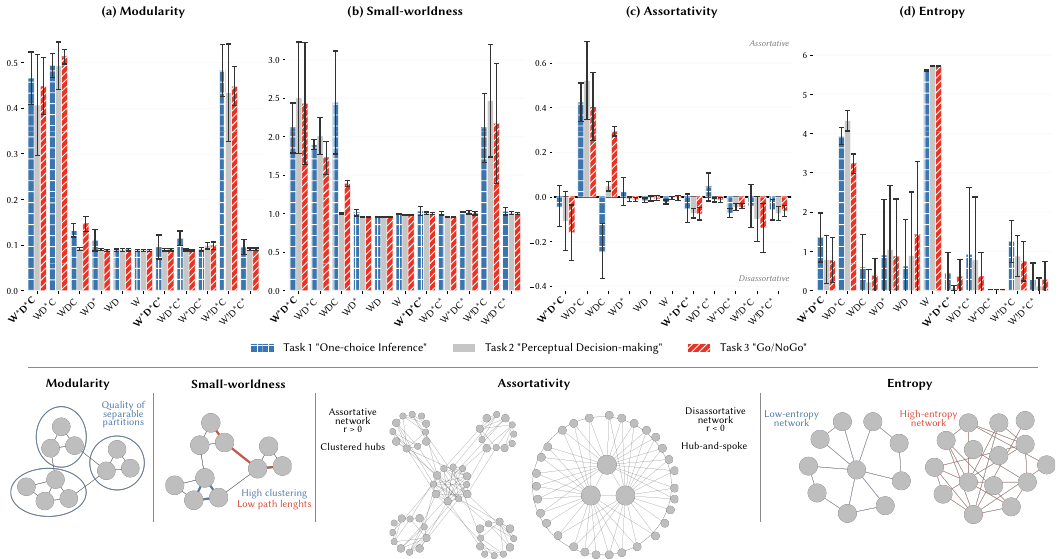}
		\caption{\label{fig:network-res}
			\textbf{Cortical priors reshape recurrent topology toward more structured regimes.} Topological properties of the learned recurrent weight matrices across model variants, evaluated using graph-theoretic metrics that capture complementary aspects of network organization. \textbf{(a) Modularity.} Modularity quantifies the extent to which the network partitions into densely connected communities with relatively sparse inter-community links. Variants combining biologically informed initialization, real spatial embedding, and direct communicability regularization generally exhibited the strongest modular structure, whereas EMD-based and more weakly constrained variants remained substantially less modular. \textbf{(b) Small-worldness.} Small-worldness measures the coexistence of strong local clustering and short global path lengths relative to appropriate random controls. Direct communicability regularization, particularly when paired with functional initialization and real coordinates, promoted robust small-world organization, whereas several \textit{C*} variants remained near random-like regimes despite learning accurate task solutions. \textbf{(c) Assortativity.} Assortativity captures whether highly connected nodes preferentially connect to other highly connected nodes (positive assortativity) or instead form hub--periphery structure (negative assortativity). Some variants, such as \textit{WD*C} and in part \textit{WDC}, developed positively assortative, hub-rich organization, whereas functionally initialized variants more often shifted toward disassortative structure, indicating that different cortical priors favor different balances between integration and distribution of computational load. \textbf{(d) Entropy.} Entropy summarizes the degree of disorder in the learned recurrent weight organization, with lower values indicating more structured, less random configurations. Function-derived initialization and real spatial embedding consistently pushed the networks away from high-entropy, random-like solutions and toward lower-entropy regimes. Notably, some variants such as \textit{W*D*C*} achieved low-entropy connectivity without comparably strong modularity or small-worldness, showing that sparse, structured recurrent organization and canonical modular small-world topology are related but distinct outcomes of learning under cortical priors.}
	\end{figure*}

	Beyond task performance, we asked whether cortical constraints also shape the organization of the learned recurrent networks. Across model variants, biologically grounded constraints produced systematic shifts in emergent topology, yielding changes in entropy, modularity, assortativity, and small-worldness that were not seen in minimally constrained models (Fig.~\ref{fig:network-res}). Importantly, these topological differences did not simply mirror performance; rather, they revealed how distinct cortical priors steer learning toward different organizational regimes.
	
	\paragraph{Entropy of weight organization.}
	Entropy analysis revealed three distinct regimes of weight structure. High-entropy variants (\textit{W} and \textit{WD*C}; entropy $\sim 3$--$6$) exhibited nearly uniform weight distributions characteristic of minimally organized connectivity and showed poor performance on Tasks 2 and 3. Intermediate-entropy variants (\textit{W*D*C}, \textit{WD*}, \textit{WD}, \textit{WD*C*}, and \textit{W!D*C}; entropy $\sim 0.6$--$1.5$) showed partially organized but still heterogeneous connectivity patterns. Low-entropy variants (\textit{WDC}, \textit{W*D*C*}, \textit{W*DC*}, and \textit{W!D*C*}; entropy $\sim 0.02$--$0.6$) exhibited highly structured, specialized connectivity. Overall, incorporating neuronal constraint components produced markedly lower-entropy organization, whereas the baseline \textit{W} model remained closer to a uniform, random-like weight regime (Fig.\ref{fig:network-res}, Entropy).
	
	\paragraph{Modularity.}
	Modularity provided a more specific view of how structured connectivity emerged across model variants. Variants such as \textit{W*D*C}, \textit{WD*C}, and \textit{W!D*C} developed the strongest community structure ($Q \sim 0.4$--$0.5$), indicating the emergence of distinct functional subnetworks. In contrast, models lacking functional priors, spatial embedding, or both remained only weakly modular ($Q \approx 0.1$). Notably, the EMD-based variants, particularly \textit{W*D*C*} and \textit{WD*C*}, achieved strong task performance and low entropy without developing comparably strong modularity. This dissociation is important because it shows that sparse, structured connectivity and modular organization are related but not identical outcomes, and that different communicability formulations can steer recurrent learning toward different topological endpoints (Fig.\ref{fig:network-res}, Modularity).
	
	\paragraph{Small-worldness.}
	Small-worldness captured an additional distinction between biologically constrained and weaker-control variants.  Models that combined functional priors, spatial embedding, and direct communicability regularization---including \textit{W*D*C}, \textit{WD*C}, \textit{WDC}, and \textit{W!D*C}---showed robust small-world structure ($\sigma \approx 1.5$--$2.5$), reflecting the coexistence of high local clustering and short global path lengths. In contrast, the EMD-based \textit{C*} variants and the more weakly constrained baselines remained near $\sigma \approx 1$, consistent with weak or random-like organization (Fig.\ref{fig:network-res}, Small-worldness). Thus, the same cortical priors that improve learning also favor a topology that balances local specialization with efficient global communication, a hallmark of biological networks.

	\paragraph{Assortativity.}
	Assortativity further differentiated the topological regimes. \textit{WD*C} variant showed consistently positive assortativity across tasks ($r \approx 0.4$--$0.5$), and \textit{WDC} was also positively assortative in Tasks 2 and 3, indicating a tendency for high-degree nodes to interconnect and form more hub-rich integrative structure. In contrast, functionally initialized variants---including \textit{W*D*C}, \textit{W*D*C*}, \textit{W*DC*}, \textit{W!D*C}, and \textit{W!D*C*}---were disassortative ($r<0$), consistent with hub--periphery organization that favors segregation and modular structure. The remaining variants stayed near zero ($r \approx 0$), indicating little systematic degree-based preference (Fig.\ref{fig:network-res}, Assortativity). These results show that cortical priors do not impose a single canonical topology. Rather, they bias recurrent learning toward distinct trade-offs between integration, segregation, and the distribution of computational load.
	
	Taken together, these findings show that cortical geometry, wiring, and function shape not only whether recurrent networks learn successfully, but also the form of the solutions they discover. The same multimodal priors that improve task performance also guide the emergence of structured connectivity regimes associated with biological organization.
	
	\subsection{\label{sec:addrobust} Additional robustness analyses support the generality of cortical priors}
	
	We performed several additional analyses to test whether the effects of biological initialization depended on a particular construction of the weight prior or on a specific sampled cortical field.
	
	First, to assess whether the exact values in the biologically initialized weight matrix carried meaningful structure beyond its overall sparsity and marginal distribution, we resampled the nonzero entries of $W_{\mathrm{bio}}$ using its empirical cumulative distribution function (ECDF) and reassigned them to their original support. These resampled initializations showed a clear decline in performance in the robustness analyses, indicating that the original biologically derived weight values contain task-relevant structure that is not preserved by distribution-matched resampling alone.
	
	Second, replacing the correlation matrix with the precision matrix when constructing the biological initialization yielded comparable performance across tasks. Because the precision matrix isolates direct functional dependencies by conditioning on all other neurons \cite{Liegeois2020funccon,Das2017precisionmatrix}, this result suggests that both direct and indirect functional relationships provide informative inductive structure for recurrent learning.
	
	Finally, robustness analyses across other session--scan--field combinations revealed consistent performance trends and emergent topological patterns despite differences in the number of nodes and sampled cortical region. This cross-field consistency suggests that the organizational principles captured by functional initialization and spatial embedding are not idiosyncratic to a single MICrONS field, but instead reflect more general cortical constraints that operate across scales and sampled regions.

	\begin{table*}
		\begin{ruledtabular}
			\begin{tabular}{lcccc}
				Model Name & Bio Weight Init & Spatial Embedding & Communicability & Regularization \\ \hline
				\textit{a.} \textit{W*D*C}   & Yes (W*)  & Real (D*) & Yes (C) & $\lambda \, \lVert W^* \odot D^* \odot C \rVert$ \\
				\textit{b.} \textit{WD*C}    & No   & Real (D*) & Yes (C) & $\lambda \, \lVert W \odot D^* \odot C \rVert$ \\
				\textit{c.} \textit{WDC }    & No   & Grid (D)  & Yes (C) & $\lambda \, \lVert W \odot D \odot C \rVert$ \\
				\textit{d.} \textit{WD*}     & No   & Real (D*) & No      & $\lambda \, \lVert W \odot D^* \rVert$ \\
				\textit{e.} \textit{WD}      & No   & Grid (D)  & No      & $\lambda \, \lVert W \odot D \rVert$ \\
				\textit{f.} \textit{W\textsubscript{(Simple RNN)}}       & No  & None      & No      & $\lambda \, \lVert W \rVert$ \\
				\hline
				\multicolumn{5}{r}{EMD (C*)} \\
				\textit{g.} \textit{W*D*C*}  & Yes (W*) & Real (D*) & EMD (C*) & $\lambda \, \lVert W^* \odot D^* \rVert + \lambda_{\text{EMD}} \cdot \text{EMD}(C_{\text{emp}}, C_{\text{art}})$ \\
				\textit{h.} \textit{WD*C*}   & No   & Real (D*) & EMD (C*) & $\lambda \, \lVert W \odot D^* \rVert + \lambda_{\text{EMD}} \cdot \text{EMD}(C_{\text{emp}}, C_{\text{art}})$ \\
				\textit{i.} \textit{W*DC*}   & Yes (W*) & Grid (D)  & EMD (C*) & $\lambda \, \lVert W^* \odot D \rVert + \lambda_{\text{EMD}} \cdot \text{EMD}(C_{\text{emp}}, C_{\text{art}})$ \\
				\hline
				\multicolumn{5}{r}{Permuted (W!)} \\
				\textit{j.} \textit{W!D*C}   & Yes (W!)  & Real (D*)  & Yes (C) & $\lambda \, \lVert W^! \odot D^* \odot C \rVert$ \\
				\textit{k.} \textit{W!D*C*}   & Yes (W!)  & Real (D*)  & EMD (C*) & $\lambda \, \lVert W^! \odot D \rVert + \lambda_{\text{EMD}} \cdot \text{EMD}(C_{\text{emp}}, C_{\text{art}})$
			\end{tabular}
		\end{ruledtabular}
		\caption{\label{tab:model_table}\textbf{Model configurations incorporating combinations of biological constraints.} \textit{W*} denotes biologically informed weight initialization derived from functional connectivity. \textit{W!} is a permutation control derived from \textit{W*} that preserves the empirical distribution of initial weight values while disrupting their original structured assignment across neuron pairs, thereby testing whether performance depends on biological structure rather than distribution alone. \textit{D*} denotes the use of real MICrONS neuronal coordinates, whereas \textit{D} uses an artificial grid as a spatial control. \textit{C} applies direct communicability-based regularization, while \textit{C*} applies an alternative Earth Mover's Distance (EMD)-based regularization over communicability distributions. Together, these components isolate the contributions of cortical function, geometry, and communication topology to recurrent learning.}
	\end{table*}

	\begin{table*}
		\begin{ruledtabular}
			\begin{tabular}{lccc}
				Model & Task 1 ``One-Choice Inference'' & Task 2 ``Perceptual Decision-Making'' & Task 3 ``Go/NoGo'' \\ \hline
				\textit{\textbf{W*D*C}}   & \textbf{0.917} (0.853, 0.981) & \textbf{0.865} (0.824, 0.906) & \textbf{0.948} (0.878, 1.000) \\
				\textit{WD*C}    & 0.628 (0.626, 0.631) & 0.812 (0.747, 0.876) & 0.939 (0.869, 1.000) \\
				\textit{WDC}     & 0.624 (0.622, 0.626) & $\approx$ 0.50 & $\approx$ 0.50 \\
				\textit{WD*}     & 0.787 (0.700, 0.874) & $\approx$ 0.50 & $\approx$ 0.50 \\
				\textit{WD}      & 0.628 (0.625, 0.630) & $\approx$ 0.50 & $\approx$ 0.50 \\
				\textit{W\textsubscript{(Simple RNN)}}       & 0.625 (0.623, 0.627) & $\approx$ 0.50 & $\approx$ 0.50 \\
				\hline
				\textit{\textbf{W*D*C*}}  & \textbf{0.985} (0.954, 1.000) & \textbf{0.880} (0.879, 0.882) & \textbf{0.951} (0.881, 1.000) \\
				\textit{WD*C*}   & 0.758 (0.672, 0.843) & $\approx$ 0.50 & $\approx$ 0.50 \\
				\textit{W*DC*}   & 0.963 (0.917, 1.000) & 0.862 (0.822, 0.901) & 0.927 (0.843, 1.000) \\
				\hline
				\textit{W!D*C}   & 0.950 (0.894, 1.000) & 0.866 (0.826, 0.906) & 0.951 (0.880, 1.000) \\
				\textit{W!D*C*}  & 0.963 (0.910, 1.000) & 0.847 (0.798, 0.897) & 0.951 (0.881, 1.000)
			\end{tabular}
		\end{ruledtabular}
		\caption{\label{tab:acc_table}\textbf{Accuracy of model variants across the three tasks, with $95\%$ confidence intervals.} Neuronal constraints were derived from MICrONS \underline{6}, scan \underline{6}, field \underline{2}. Results are based on 20 runs over 10 epochs using 312 nodes.}
	\end{table*}
	
	\begin{table}
		\begin{ruledtabular}
			\begin{tabular}{lccc}
				Model & Task 1 & Task 2 & Task 3 \\ \hline
				\textit{\textbf{W*}D*C}   & 0.910 & 0.834 & 0.950 \\
				\textit{WD*C}    & $\approx$ 0.25 & $\approx$ 0.50 & $\approx$ 0.50 \\
				\textit{WDC}     & $\approx$ 0.25 & $\approx$ 0.50 & $\approx$ 0.50 \\
				\textit{WD*}     & $\approx$ 0.25 & $\approx$ 0.50 & $\approx$ 0.50 \\
				\textit{WD}      & $\approx$ 0.25 & $\approx$ 0.50 & $\approx$ 0.50 \\
				\textit{W}       & $\approx$ 0.25 & $\approx$ 0.50 & $\approx$ 0.50 \\
				\hline
				\textit{\textbf{W*}D*C*}  & 1.000 & 0.838 & 1.000 \\
				\textit{WD*C*}   & $\approx$ 0.25 & $\approx$ 0.50 & $\approx$ 0.50 \\
				\textit{\textbf{W*}DC*}   & 0.945 & 0.845 & 0.950 \\
				\hline
				\textit{\textbf{W!}D*C}   & 0.804 & 0.854 & 0.975 \\
				\textit{\textbf{W!}D*C*}  & 0.988 & 0.877 & 0.924
			\end{tabular}
		\end{ruledtabular}
		\caption{\label{tab:acc_table_sign}\textbf{Task accuracy under positive-only recurrent weights.} Only functionally initialized variants (\textit{W*}, \textit{W!}) retained high performance, whereas randomly initialized variants collapsed toward chance.}
	\end{table}

	\section{\label{sec:discussion} Discussion}
	
	\emph{Cortical priors improve recurrent learning rather than merely making the models more biologically interpretable:} In this study we addressed a foundational question: which aspects of cortical organization provide useful inductive structure for recurrent computation? Using data released through the Machine Intelligence from Cortical Networks (MICrONS) program, we constructed recurrent neural networks constrained by measured cortical geometry, wiring, and function, and systematically compared the contribution of each component across three cognitive tasks. The central result is clear: biologically grounded cortical priors did not simply make recurrent networks more ``brain-like''; they made them better learners. Across tasks, networks informed by cortical structure and function outperformed unconstrained and partially constrained variants while simultaneously converging toward topologies associated with biological circuits. These findings provide direct empirical support for long‑standing proposals that cortical microcircuits are organized around efficiency principles  \cite{bae2025functionalconnectomicsspanning,achterberg2023spatiallyembeddedrecurrent,pulvermuller2021biologicalconstraintson,Vanderhaeghen2023evo,Chklovskii2002wiringOptim}.
	
	\emph{The different cortical priors do not contribute equally.} Functional weight initialization provided the largest and most general performance gain, real spatial embedding contributed a robust additional benefit across tasks, and communicability exerted a more selective effect that depended on architectural context. This hierarchy matters scientifically because it helps isolate which features of cortical organization are most consequential for recurrent learning. It also matters for better design of artificial systems because it shows that meaningful gains can be achieved not by increasing architectural complexity, but by reshaping the optimization landscape with biologically grounded priors. In that sense, the present work is not only about improving biological realism; it is about identifying which aspects of cortical organization act as useful computational constraints —finding evidence for such claims \cite{achterberg2023spatiallyembeddedrecurrent,sheeran2024spatialembeddingpromotes,pulvermuller2021biologicalconstraintson}. The formal nonparametric decomposition of these effects is reported in Supplementary Results and Supplementary Tables~\ref{tab:S5_mw_pairwise_W}--\ref{tab:S7_mw_pairwise_C}.
	
	\emph{The strongest models suggest a structured synergy between cortical geometry and cortical function.} The best-performing variants, \textit{W*D*C} and \textit{W*D*C*}, combined function-derived initialization with real neuronal coordinates and communicability-aware regularization. Their superiority over partially constrained variants indicates that cortical geometry and function interact synergistically rather than redundantly. At the same time, the ablation analyses show that this synergy is not uniform: the dominant contribution comes from functional initialization, whereas spatial embedding and communicability refine the space of learned solutions. Detailed omnibus and pairwise comparisons for these ablation effects are provided in Supplementary Results. This result is important because it suggests that the value of cortical data lies not only in anatomical wiring diagrams, but also in activity-derived relationships that encode how neurons participate in shared computation. The MICrONS resource is especially powerful in this respect because it co-registers positions, connectivity, and activity in the same neuronal population \cite{bae2025functionalconnectomicsspanning}. MICrONS also provides access to general wiring rules \cite{Ding2025WiringRule}, and connectomic datasets more broadly offer principled constraints on branching patterns and wiring motifs that can be leveraged in artificial architectures \cite{Dorkenwald2024connectomeWire,Budd2012wiringCortex,Cuntz2010GrowingWire}.
	
	\emph{Under positive‑weight constraints, bio‑inspired initialization provides a substantive inductive bias for recurrent learning.} When recurrence was restricted to positive weights, randomly initialized models collapsed to near-chance performance, whereas biologically initialized models retained high accuracy. Function-derived initialization therefore did more than improve average performance under standard training; it stabilized learning when recurrent optimization became substantially more difficult. This is precisely what one expects from a useful inductive bias: it narrows the space of candidate solutions toward regions that are both easier to optimize and more computationally effective. The comparable performance of \textit{W!} and \textit{W*} further shows that this benefit depends primarily on the statistical structure of the biological initialization rather than on a strict one-to-one mapping between neuronal position and initial weight identity. At the same time, the degradation observed after ECDF resampling indicates that not all biologically matched distributions are equivalent; the specific pattern of biologically derived weight values retains information that is lost under distribution-preserving randomization. These observations fit well with recent work showing that sign constraints and recurrent spectra can strongly shape trainability in biologically constrained RNNs \cite{song2016trainingexcitatoryinhibitoryrecurrent,li2023learningbetterwith,balwani2025constructingbiologicallyconstrained}.
	
	\emph{Cortical priors altered the solution class reached by learning, with their topology shifting convergence from random‑like to structured, low‑entropy recurrent regimes.} Model variants constrained by cortical structure and function did not simply achieve higher accuracy; they converged to different recurrent regimes. Entropy analysis revealed a progression from high-entropy, random-like configurations to low-entropy, highly structured weight organizations. Functionally initialized and spatially grounded models were consistently pushed away from the random regime and toward more specialized connectivity. This supports the view that cortical priors reduce the degeneracy of recurrent optimization and narrow the set of feasible solutions toward more organized weight structures \cite{mcallister2026nonrandombrainconnectome,Reimann2026nonrandomArchitect}. The inverse relationship between entropy and task performance further suggests that improved learning in these models is associated with concentrating synaptic resources on a smaller set of more informative connections. In this respect, the present results align with recent observations that spatially embedded recurrent networks preferentially converge toward low-entropy structured solutions \cite{sheeran2024spatialembeddingpromotes,beiran2025predictionofneural}. This echoes the principle that small‑world topologies support efficient information transmission: networks that combine local clustering with short global paths maximize communication efficiency and information throughput \cite{Aprile2022smallworldOptim,Latora2001smallworldEfficient}. The structured, low‑entropy regimes reached by our cortical‑prior models reflect this same efficiency‑driven bias.
	
	\emph{Cortical priors introduced an additional layer of modular and small‑world organization, though this structure did not take a single canonical form.} Direct communicability regularization, particularly when combined with biological initialization and real spatial embedding, consistently promoted modular and small-world organization. These topologies are of interest because they balance local specialization with efficient global and local communication, a combination widely associated with cortical network organization at both macro and microscales \cite{meunier2010modularandhierarchically,bullmore2012theeconomyof,gallos2012asmallworld,petersen2015brainnetworksand,Chen2017primateconnectome}. This pattern is also consistent with prior work showing that wiring-economical constraints can promote modularity, clustering, and other brain-like topological features while preserving or even improving task performance \cite{Chen2006wiringoptimization,Varshney2011celegansconnectome,Cherniak1992localoptimizationarbor}. In that sense, the modular and small-world regimes observed here are not incidental byproducts of sparsification, but expected consequences of optimizing recurrent computation under spatial cost \cite{Zhang2025wiringeconomy,Clune2013evomodularity,Masuda2004smallworldSynch}. Yet one of the more interesting findings here is that these properties were not obligatorily linked to task success. The \textit{W*D*C*} variant achieved accuracy comparable to \textit{W*D*C} and developed low-entropy connectivity, but showed substantially weaker modularity and near-random small-worldness. This dissociation indicates that sparse, structured connectivity and modular small-world organization are related but distinct outcomes. Different ways of enforcing communicability can steer recurrent learning toward different organizational solutions that remain functionally adequate. In other words, cortical priors constrain the family of solutions, but do not impose a single topological endpoint.
	
	\emph{Assortativity further suggests that different priors favor different trade-offs between integration and distribution of computational load.} Some variants, notably \textit{WD*C} and in part \textit{WDC}, developed positively assortative, hub-rich organization, whereas the functionally initialized models tended toward more disassortative hub--periphery structure. These are not minor graph-theoretic details. They imply that different cortical priors favor different balances between integration, segregation, and robustness. The tendency of functionally initialized models toward disassortative organization suggests that functional priors may bias recurrent networks toward more distributed architectures, rather than toward densely interconnected hub cores. More broadly, these results indicate that cortical structure--function priors shape not only whether a model learns, but also what kind of recurrent solution it discovers \cite{mcallister2026nonrandombrainconnectome,Varshney2011celegansconnectome,Chen2006wiringoptimization}.
	
	\emph{Methodologically, by moving beyond stylized constraints and grounding recurrent models in measured cortical organization, we found that distinct structural and functional biases emerge.} Prior studies showed that abstract spatial and communication constraints can improve performance and induce more brain-like organization in recurrent networks \cite{achterberg2023spatiallyembeddedrecurrent,sheeran2024spatialembeddingpromotes}. The present work moves beyond stylized embeddings by grounding recurrent models in measured cortical geometry, wiring, and function. MICrONS is particularly important in this regard because it combines dense calcium imaging with co-registered high-resolution electron microscopy from the same animal and cortical substrate, thereby linking neuronal positions, anatomical connectivity, and functional relationships in one multimodal resource \cite{bae2025functionalconnectomicsspanning}. In this sense, our study helps bridge structural connectomics and artificial recurrent computation. It also sits naturally alongside recent efforts to extend seRNN-style approaches toward connectome-constrained formulations \cite{rovny2024connectomeconstrainedspatiallyembedded}.
	
	\emph{Anatomical structure shaped recurrent dynamics but did not uniquely specify them, with functional information providing the decisive constraint.} This multimodal dependence reflects the broader principle that structure alone often does not uniquely determine dynamics. Recent connectome‑constrained modeling work has shown that anatomical connectivity can powerfully constrain neural computation, but may still admit multiple dynamical realizations unless additional physiological or activity‑based information is included \cite{lappalainen2024connectomeconstrainednetworkspredict,beiran2025predictionofneural,park2013structuralandfunctional,lynn2019thephysicsof}. Our results are consistent with that caution. Anatomical geometry and wiring mattered, but the strongest effect came from function-derived initialization, and the similar performance of correlation-based and precision-based initializations suggests that both direct and indirect functional dependencies contain useful information for constraining learning. The present study therefore uses a multimodal cortical resource not only to constrain models, but also to ask a more basic scientific question: what kinds of biological information are most computationally informative?
	
	\emph{Biological priors improved performance and robustness in ways directly relevant to the design of artificial systems beyond neuroscience.} In the present framework, biological priors improved performance and robustness without adding layers, parameters, or specialized modules. Instead, they changed how learning began and how optimization unfolded. This is a useful lesson for machine learning: better recurrent computation need not always come from larger models or more elaborate architectures; it can also come from better inductive structure. For domains in which data efficiency, robustness, and interpretable internal organization are especially valuable, cortical structure--function priors may offer a principled alternative to purely generic initialization and regularization strategies \cite{pulvermuller2021biologicalconstraintson,marblestone2016towardanintegration}. Cortical structural motifs can likewise serve as inductive priors for building such systems, offering reusable wiring patterns that inform artificial architectures \cite{Matelsky2021motif,johnson2023exploitinglargeneuroimagingdatasets}.
	
	\emph{The present findings come with several scope‑defining limitations.} First, MICrONS captures cortical structure and function under a specific experimental regime, and the functional relationships extracted here may depend on stimulus context, imaging depth, and behavioral state \cite{bae2025functionalconnectomicsspanning}. Second, the present analysis focused on excitatory neurons, leaving out inhibitory cell types that are central to cortical dynamics and circuit motif structure \cite{schneidermizell2024celltypespecificinhibitorycircuitry}. The diversity of inhibitory morphoelectric types \cite{Yanez2026morphoelectricInhibitory} itself can be used as a prior in the design of bio-inspired RNNs.  Third, the functional measures used here capture statistical dependence rather than causal interaction. Finally, the task battery, while appropriate for controlled comparison, remains  relatively simple compared with the richness of naturalistic cognition. None of these limitations undermines the main result, but they do define its scope: the study identifies cortical priors that improve recurrent learning in a controlled setting, rather than exhaustively capturing cortical computation in full biological detail.
	
	\emph{A key advantage of the bio‑inspired construction is the robustness of its performance across different instantiations of the prior.} The robustness analyses nevertheless suggest that the effects are not idiosyncratic to a single construction. Performance remained comparable when correlation-based weight initialization was replaced by precision-based initialization, indicating that multiple function-derived statistics can serve as informative priors. This is informative because the precision matrix emphasizes conditional, comparatively more direct statistical dependencies after removing variance shared through the rest of the network \cite{Das2017precisionmatrix,Liegeois2020funccon}. The comparable performance of precision- and correlation-based priors therefore suggests that recurrent learning can be guided both by broad co-activity structure and by a sparser scaffold of putatively direct functional relationships \cite{Liu2025benchmarkconnectivity}. Likewise, analyses across other session--scan--field combinations preserved the main performance and topology trends despite changes in neuron number and sampled cortical region. Together, these controls suggest that the computational value of cortical geometry and function is not confined to a single field or a single functional estimator, but reflects more general organizational structure in the underlying data—an organizational principle anticipated in prior work \cite{bae2025functionalconnectomicsspanning} and now demonstrated directly by the present results.
	
	\emph{More broadly, the framework introduced here provides a platform for reverse-engineering cortical computation.} By independently manipulating geometry, wiring, and functional priors, this model family makes it possible to test which biological ingredients are necessary for which computational outcomes. This reciprocity may run in both directions: just as cortical organization can guide artificial recurrent design, task optimization can also recover wiring regularities observed in MICrONS-like data \cite{Ding2025WiringRule}. That convergence suggests that at least some cortical connectivity motifs may reflect computational pressures that trained recurrent systems rediscover, rather than anatomical contingency alone. \emph{Within this framework, recurrent networks grounded in cortical geometry, wiring, and function learned more effectively and converged toward more structured organizational regimes than unconstrained alternatives.} Function-derived initialization exerted the strongest influence, spatial embedding added a reliable secondary benefit, and communicability shaped the topology of the learned solution. Taken together, these results show that cortical machinery can serve as a powerful inductive basis for the design of artificial learning systems and a systematic experimental framework for identifying which features of cortical organization are computationally consequential.

	\section{\label{sec:methods} Methods}
	For an overview of the methods see Fig.~\ref{fig:overview}.

	\subsection{\label{subsec:microns} Dataset}
	We used functional connectomics data from the MICrONS (Machine Intelligence from Cortical Networks) public dataset \cite{bae2025functionalconnectomicsspanning}. MICrONS provides multimodal, large-scale measurements from mouse visual cortex, including primary visual cortex and three higher visual areas. The dataset contains detailed anatomical reconstructions of more than 200,000 cells and 523 million synapses, together with two-photon calcium imaging recordings of visual responses from approximately 75,000 neurons.
	
	Among these neurons, approximately 12,000 excitatory cells have been functionally coregistered, meaning that both structural connectivity from electron microscopy and functional activity from calcium imaging are available for the same neurons. This coregistration enabled us to derive spatial, structural, and functional constraints from a common neuronal population.
	
	\subsubsection{Structural Connectivity}
	We analyzed multiple session--scan--field combinations from the MICrONS dataset (Supplementary Table~\ref{tab:S9_session_scan_fields}). Here, a \emph{session} denotes a calcium imaging experiment, a \emph{scan} denotes a specific acquisition within that session, and a \emph{field} denotes a particular cortical region imaged across scans and sessions.
	
	Structural connectivity data were accessed through the \texttt{CaveClient} interface \cite{dorkenwald_cave_2025}. We queried the \texttt{coregistration\_manual\_v4} table, which contains manually verified links between structural and functional records. From this table, we extracted \texttt{root\_id}s, soma positions, and \texttt{unit\_id}s. The \texttt{root\_id} is the unique identifier of a neuron in the proofread segmentation, the position gives the soma location in 4\texttimes4\texttimes40~nm voxels, and the \texttt{unit\_id} is the functional ROI identifier, unique within each scan.
	
	For each field, we then constructed a neuronal connectivity graph by querying the database for pre- and postsynaptic connections among the selected neurons.
	
	\subsubsection{Functional Activity}
	The MICrONS dataset also includes two-photon calcium imaging recordings from approximately 75,000 excitatory neurons spanning cortical layers 2--5 across four visual areas---VISp, VISlm, VISrl, and VISal---in a transgenic mouse expressing GCaMP6s.
	
	The visual stimulus set included both naturalistic and parametric stimuli designed to sample a broad visual feature space \cite{bae2025functionalconnectomicsspanning}. In this study, the functional measures were computed from responses to three stimulus classes used in the MICrONS imaging protocol: \textit{Clip}, \textit{Monet2}, and \textit{Trippy}. \textit{Clip} consists of natural video segments drawn from cinematic footage, sports videos, and rendered first-person virtual environments, providing complex real-world visual statistics across a broad feature space. \textit{Monet2} is a global directional parametric stimulus constructed from spatially and temporally smoothed noise with coherent orientation and motion, designed to probe tuning to global direction and orientation. \textit{Trippy} is a local directional parametric stimulus generated from transformed smooth noise, designed to probe local visual features including orientation, direction, and spatial and temporal frequency.
	
	These recordings were acquired \textit{in vivo} across 14 densely sampled scans while the animal viewed a range of visual stimuli and behavioral variables were recorded in parallel.
	
	To access the functional data, we cloned the \textit{\href{https://github.com/datajoint/microns_phase3_nda}{microns\_phase3\_nda}} repository, which provides access to the MICrONS Phase 3 functional dataset. Using the \texttt{root\_id}s and \texttt{unit\_id}s of neurons with known structural connectivity, we queried deconvolved spike data derived from the calcium imaging traces.
	
	\subsection{Functional Calculations\label{subsec:functional_calc}}
	Structural connectivity defines potential synaptic relationships, but it does not by itself capture the dynamic interactions expressed during activity. To estimate functional relationships among neurons, we computed three complementary measures from the Ca imaging data: the Pearson correlation coefficient \cite{Pearson_Corr_1895}, the Spike Time Tiling Coefficient (STTC) \cite{cutts_sttc_2014}, and the inverse covariance (precision) matrix \cite{Das2017precisionmatrix,Dawid1979conditionalindepence}.
	
	These measures were used to incorporate empirically observed functional structure into the recurrent models, alongside anatomical and spatial constraints.
	
	\subsubsection{Correlation Coefficient}
	For each scanned field, we computed Pearson correlation coefficients \cite{Pearson_Corr_1895} from the functional activity of all neurons.
	
	Spike trains were first discretized into bins. Pairwise correlations were then computed for all pairs among the \(N\) binned spike trains, yielding a symmetric \(N \times N\) matrix in which each element \(C[i,j]\) represents the correlation between neurons \(i\) and \(j\).
	
	Let \(b_i\) and \(b_j\) denote the binned spike trains of neurons \(i\) and \(j\), and let \(\mu_i\) and \(\mu_j\) denote their corresponding means. The correlation coefficient was calculated as
	
	\begin{equation}
		C[i, j] = \frac{\langle b_i - \mu_i,\, b_j - \mu_j \rangle}{\sqrt{\langle b_i - \mu_i,\, b_i - \mu_i \rangle \cdot \langle b_j - \mu_j,\, b_j - \mu_j \rangle}},
	\end{equation}
	
	where \(\langle \cdot, \cdot \rangle\) denotes the scalar product.
	
	\subsubsection{STTC (Spike Time Tiling Coefficient)}
	We also computed the Spike Time Tiling Coefficient (STTC), introduced by Cutts and Eglen \cite{cutts_sttc_2014}, as a complementary pairwise measure of dependence between spike trains.
	
	Compared with Pearson correlation, STTC is less sensitive to firing-rate confounds, distinguishes uncorrelated from anti-correlated activity, does not treat silent periods as informative evidence of correlation, and is more sensitive to temporal spike relationships.
	
	For each neuronal pair \(A\) and \(B\), STTC was computed as
	
	\begin{equation}
		\text{STTC} = \frac{1}{2}\left(\frac{P_A - T_B}{1 - P_A T_B} + \frac{P_B - T_A}{1 - P_B T_A}\right),
	\end{equation}
	
	where \(T_A\) is the proportion of the recording that lies within \(\pm \Delta t\) of any spike from neuron \(A\), and \(T_B\) is defined analogously for neuron \(B\). Likewise, \(P_A\) is the proportion of spikes from neuron \(A\) that fall within \(\pm \Delta t\) of any spike from neuron \(B\), and \(P_B\) is defined analogously.
	
	\subsubsection{Precision Matrix}
	Finally, we computed the precision matrix, which isolates direct statistical dependencies by removing correlations mediated by other neurons.
	
	Whereas the correlation matrix reflects both direct and indirect dependencies, each entry \(\Theta[i,j]\) of the precision matrix reflects the conditional dependence between neurons \(i\) and \(j\) given the activity of all other neurons. Thus, unlike the correlation matrix, the precision matrix excludes all indirect influences, retaining only pairwise interactions that cannot be explained by any other neuron \cite{Liegeois2020funccon, Das2017precisionmatrix}. We therefore used the precision matrix as an alternative functional prior to assess whether direct dependencies alone were sufficient to guide recurrent learning.
	
	The precision matrix \(\Theta\) was obtained as the inverse of the covariance matrix \(\Sigma\) \cite{Dawid1979conditionalindepence}:
	
	\begin{equation}
		\Theta = \Sigma^{-1}.
	\end{equation}
	
	Let $b_i$ and $b_j$ denote the binned spike trains for neurons $i$ and $j$, and let $\mu_i$ and $\mu_j$ be their respective mean values. The covariance matrix was defined as
	
	\begin{equation}
		\Sigma[i,j] = \langle b_i - \mu_i,\, b_j - \mu_j \rangle,
	\end{equation}
	
	where $\langle \cdot, \cdot \rangle$ denotes the scalar (dot) product of two vectors. 
	
	\subsection{\label{subsec:sernns} Anatomically and Functionally Constrained Recurrent Neural Networks}
	Our modeling framework builds on the concept of spatially embedded recurrent neural networks (seRNNs) \cite{achterberg2023spatiallyembeddedrecurrent}, but replacing their abstract spatial embeddings with constraints derived directly from cortical data. We used this framework to test how cortical geometry, wiring, and function influence recurrent learning.
	
	We defined eleven RNN variants that incorporated all, some, or none of the biological constraints under study. Across these models, recurrent weights could be initialized from functional data, recurrent units could be embedded either in real neuronal coordinates or in an artificial spatial grid, and regularization could include either direct communicability or an Earth Mover’s Distance (EMD) term enforcing similarity between empirical and artificial communicability distributions. All models were trained on the same three cognitive tasks and evaluated using the same performance and graph-theoretic metrics.
	
	In the original seRNN formulation \cite{achterberg2023spatiallyembeddedrecurrent}, each recurrent unit is placed on a regular three-dimensional grid and pairwise Euclidean distances are used to construct a distance matrix \(D\). To incorporate network topology, a communicability matrix \(C\) is also computed, capturing the extent to which information can propagate between nodes through walks of all lengths. Following \cite{achterberg2023spatiallyembeddedrecurrent}, communicability is given by
	
	\begin{equation}
		\label{eq:5}
		C = e^{S^{-1/2} W S^{-1/2}},
	\end{equation}
	
	where \(W\) is the adjacency matrix and \(S\) is the diagonal matrix of node strengths. Diagonal entries were set to zero to exclude self-communicability.
	
	The spatial constraint is incorporated through an \(\ell_1\) regularization term that modulates each recurrent weight $W_{ij}$ by both distance  $D_{ij}$ and communicability $C_{ij}$:
	
	\begin{equation}
		\mathcal{L} = \mathcal{L}_{\text{task}} + \lambda \, \lVert W \odot D \odot C \rVert,
	\end{equation}
	
	where \(\mathcal{L}_{\text{task}}\) is the task loss, \(\lambda\) is a regularization coefficient, and \(\odot\) denotes element-wise multiplication. This term encourages the model to form spatially economical yet communicatively efficient recurrent structure.
	
	\subsubsection{Weight Initialization}
	To incorporate biologically informed initial conditions, we initialized recurrent weights using functional relationships derived from neuronal activity. Specifically, we used the correlation, STTC, and precision matrices computed from the MICrONS data. These matrices were loaded from preprocessed files and normalized with min--max scaling so that their contributions were on comparable numerical scales.
	
	Base recurrent weights were sampled from a log-normal distribution with parameters \(\mu = -0.5\) and \(\sigma = 0.5\), chosen to reflect biologically plausible heavy-tailed synaptic weight distributions \cite{song_highly_2005, loewenstein_multiplicative_2011}. The sampled weights were then modulated element‑wise by the normalized functional matrices, embedding functional dependencies directly into the initial connectivity. In the primary initialization, the log-normal matrix was combined with the correlation and STTC matrices:
	
	\begin{equation}
		\label{eq:7}
		W_{\text{bio}} = W_{\text{lognormal}} \odot \text{Corr} \odot \text{STTC},
	\end{equation}
	
	where \(W_{\text{bio}}\) is the biologically informed recurrent initial weight matrix, \(W_{\text{lognormal}}\) is the sampled log-normal weight matrix, and \(\text{Corr}\) and \(\text{STTC}\) are the normalized Pearson correlation coefficient and normalized Spike Time Tiling Coefficient matrices, respectively. $\odot$ denotes the element-wise product.
	
	To test whether direct dependencies were sufficient, we also evaluated an alternative version in which the correlation matrix was replaced by the precision matrix.
	
	After construction, biologically initialized weights were rescaled to a fixed mean of 0.1, and a minimum threshold of 0.01 was applied to avoid near-zero connections. To promote stable recurrent dynamics, each matrix was rescaled to a spectral radius of 0.95 using its largest eigenvalue. \cite{Pascanu_difficulty_2013}. Model variants without biological weight initialization used \texttt{Orthogonal} initialization instead.
	
	\subsubsection{Model Architecture and Variant Definitions}
	All eleven model variants shared the same core RNN architecture and differed only in how biological constraints were imposed. The hidden state and output were defined as
	
	\begin{equation}
		h_t = \mathrm{ReLU}(W_x x_t + W_h h_{t-1} + b_h),
	\end{equation}
	\begin{equation}
		\hat{y}_t = \sigma(W_y h_t + b_y),
	\end{equation}
	
	where \(x_t\) is the input at time \(t\), \(h_t\) is the hidden state, \(W_x\) and \(W_h\) are the input and recurrent weight matrices, \(b_h\) is the hidden bias, \(W_y\) and \(b_y\) are the output weights and bias, and \(\sigma(\cdot)\) denotes the softmax nonlinearity.
	
	Our naming convention reflects the constraints used in each model variant. Models prefixed with \textit{W*} use biologically informed initialization defined by Eq.~\eqref{eq:7}. Models prefixed with \textit{D*} use real neuronal coordinates from MICrONS for spatial embedding. Models prefixed with \textit{C*} use an Earth Mover’s Distance (Wasserstein distance) loss between empirical and artificial communicability distributions. 
	
	Furthermore, models prefixed with \textit{W!} use the same biologically informed initialization as \textit{W*}, but with the weights permuted to break the one-to-one correspondence between initial weights and neuronal positions. 
	
	Finally, models prefixed with \textit{C} include communicability directly in the regularization term, whereas models prefixed with \textit{D} use randomly assigned grid coordinates instead of real neuronal positions.
	
	The eleven model variants were defined as follows ( For an overview of the models, see Table~\ref{tab:model_table}). \\
	
	\paragraph{W*D*C}
	Recurrent network with biological
	weight initialization, real neuronal coordinates and direct communicability regularization.
	
	\[
	\mathcal{L} = \mathcal{L}_{\text{task}} + \lambda \, \lVert W \odot D \odot C \rVert
	\]

	\paragraph{WD*C}
	Same as \textit{W*D*C}, but without biologically informed weight initialization.
	
	\paragraph{WDC}
	Network with artificial spatial coordinates and artificial direct communicability regularization, but without biologically informed initialization or real neuronal positions.
	
	\paragraph{WD*}
	Network regularized by spatial distance using real neuronal coordinates, but without communicability and without biologically informed initialization.
	
	\[
	\mathcal{L} = \mathcal{L}_{\text{task}} + \lambda \, \lVert W \odot D \rVert
	\]

	\paragraph{WD}
	Same as \textit{WD*}, but using an artificial spatial grid rather than real neuronal coordinates.
	
	\paragraph{W\textsubscript{(Simple RNN)}}
	Baseline (vanilla) recurrent network with no biological initialization and only standard \(\ell_1\) regularization on the recurrent weights.
	
	\[
	\mathcal{L} = \mathcal{L}_{\text{task}} + \lambda \, \lVert W \rVert
	\]

	\paragraph{W*D*C*}
	Network with biological weight initialization, real neuronal coordinates and an EMD penalty that matches empirical and model communicability distributions. We refer to this as the \textit{full model}.
	\[
	\mathcal{L} = \mathcal{L}_{\text{task}} 
	+ \lambda \, \lVert W \odot D \rVert 
	+ \lambda_{\text{EMD}} \cdot \mathrm{EMD}(\mathbf{C}_{\text{emp}}, \mathbf{C}_{\text{art}})
	\]

	\paragraph{WD*C*}
	Same as \textit{W*D*C*}, but without biologically informed initialization.
	
	\paragraph{W*DC*}
	Same as \textit{W*D*C*}, but using artificial grid coordinates instead of real neuronal coordinates.
	
	\paragraph{W!D*C}
	Same as \textit{W*D*C}, but with the biologically informed weights permuted.
	
	\paragraph{W!D*C*}
	Same as \textit{W*D*C*}, but with the biologically informed weights permuted.
	
	\subsubsection{Training}
	Each model variant was trained on the same three cognitive tasks to enable direct comparison across constraint conditions.
	
	All models were implemented in TensorFlow using the Keras API. The architecture consisted of a Gaussian noise layer with \(\sigma = 0.05\), a \texttt{SimpleRNN} layer with \texttt{ReLU} activation and a number of hidden units equal to the number of neurons in the corresponding MICrONS field, and a dense output layer with softmax activation.
	
	Neuronal soma coordinates were normalized independently along each spatial axis using min--max scaling. Pairwise Euclidean distances computed from these normalized coordinates therefore provided dimensionless spatial distances that were directly comparable across fields.
	
	Communicability was computed online during training from the absolute recurrent weight matrix using the degree-normalized matrix exponential in Eq.~\eqref{eq:5}, with diagonal entries removed. 
	
	For models with the EMD-based loss, the empirical communicability matrix \(\mathbf{C}_{\text{emp}}\) was computed from the weighted connectivity graph derived from MICrONS electron microscopy data using the same formulation. Both empirical and artificial communicability matrices were flattened and treated as distributions of communicability values, and the Wasserstein distance was computed between their sorted quantiles.
	
	Models with biologically informed weight initialization (\textit{W*} and \textit{W!}) used a custom initializer based on the precomputed matrix \(W_{\text{bio}}\), as defined in Equation \eqref{eq:7}. All other models used \texttt{Orthogonal} initialization. Regularization was applied according to each model definition, with \(\lambda = 0.3\) and \(\lambda_{\text{EMD}} = 0.1\). These regularization strengths were selected in pilot experiments and then held fixed across all tasks, fields, and model variants. The qualitative differences among models remained stable under moderate parameter variation.
	
	To account for variability in initialization and training, each model was trained in 20 independent simulation runs. In each simulation run, training used the Adam optimizer and categorical cross-entropy loss and was carried out for 10 epochs. In pilot runs extending training to 50 epochs, performance and topological metrics plateaued within the first 10 epochs across all variants, with no substantive improvement thereafter. For each simulation run, we recorded accuracy, loss, entropy, modularity, assortativity, and small-worldness.
	
	To evaluate robustness under sign constraints, we also trained versions of each model in which recurrent weights were constrained to be non-negative, consistent with the fact that the functional and structural data used for initialization were derived from excitatory neurons.
	
	\subsection{\label{subsec:tasks} Tasks}
	To evaluate recurrent learning under distinct temporal and cognitive demands, we trained all model variants on three widely used tasks in neuroscience and machine learning: One-Choice Inference \cite{achterberg2023spatiallyembeddedrecurrent}, Perceptual Decision-Making \cite{britten_pdm_1992}, and Go/NoGo \cite{zhang_gonogo_2019}.
	
	These tasks span different combinations of temporal integration, working memory, and context-dependent decision making. Together, they provide a broader test of the models’ ability to maintain information over delays, integrate sensory evidence over time, and generate appropriate outputs only when required.
	
	For each task, the training set contained 5,120 trials, while the validation and test sets each contained 2,560 trials. All models were trained with a batch size of 128. A full list of all task conditions is provided in Supplementary Table~\ref{tab:S8_all_tasks}.
	
	\subsubsection{One-Choice Inference}
	In the One-Choice Inference task \cite{achterberg2023spatiallyembeddedrecurrent}, the network must integrate two sequential stimuli before making a decision. Stimulus A is presented for 20 time steps, followed by a 10-step delay, after which stimulus B is presented for another 20 time steps. The network then makes a single choice at the end of the trial (Fig.~\ref{fig:tasks}, left).
	
	The task can be interpreted as a one-step navigation problem. Stimulus A specifies a goal location, and stimulus B presents possible movement directions. The correct response is the direction that would move the agent closer to the goal.
	
	Inputs were one-hot encoded with eight binary channels. The first four channels encoded the goal location, with exactly one active during the first stimulus period. The remaining four channels encoded the candidate movement directions, with two active during the second stimulus period (i.e., the choice‑relevant stimulus).
	
	\subsubsection{Perceptual Decision-Making}
	In the Perceptual Decision-Making task \cite{britten_pdm_1992}, the network must determine which of two alternatives is dominant in a noisy sensory input (Fig.~\ref{fig:tasks}, middle). The task is modeled after random-dot motion paradigms in which the subject identifies the dominant motion direction under varying coherence levels \cite{Newsome1988,Shadlen2001}.
	
	Each trial consisted of three phases: a fixation period of 1 time step, a stimulus period of 30 time steps, and a delay period of 10 time steps. During fixation and delay, the network received a constant input. During the stimulus period, it received noisy evidence favoring one of two alternatives, with coherence levels of 0, 6.4, 12.8, 25.6, or 51.2\%.
	
	Task difficulty was controlled by coherence level, where higher coherence means stronger signal-to-noise ratio. At 0\% coherence, the two alternatives are equally supported and the decision is effectively random; at 51.2\% coherence, one alternative is strongly favored. Inputs were encoded with one fixation channel and two stimulus-evidence channels. Outputs were one-hot encoded to represent the two possible choices. The network produced a single response at the end of the trial.
	
	\subsubsection{Go/NoGo}
	In the Go/NoGo task \cite{zhang_gonogo_2019}, the network had to decide whether to respond (Go) or withhold a response (No-Go) at the end of the trial. Each trial consisted of a fixation period of 5 time steps, a stimulus period of 20 time steps, a delay period of 10 time steps, and a decision period of 5 time steps (Fig.~\ref{fig:tasks}, right). Fixation and delay were represented by a constant input, whereas the stimulus period indicated whether a Go or No-Go response was required. The network responded only at the final time step.
	
	This task probes the ability to retain a stimulus across a short delay and to gate the output appropriately in time. A Go stimulus requires an action, whereas a No-Go stimulus requires suppressing action until the response period. Since the correct response is only revealed during the stimulus phase and executed after a delay, the task places demands on short-term memory and control over output timing.
	
	Inputs were one-hot encoded using three binary channels. The first channel represented fixation, and the remaining two encoded stimulus identity. Outputs were one-hot encoded with two channels corresponding to Go and No-Go responses.
	
	\subsection{\label{subsec:network_outcome} Network Outcome Assessment}
	To compare the model variants on common footing, we evaluated each model across 20 independent simulation runs and quantified both task performance (accuracy, loss and entropy) and emergent network organization (modularity, assortativity,
	and small-worldness).
	
	All graph-theoretic metrics were computed on undirected binary graphs derived from the trained recurrent weight matrices. Recurrent weights were converted to absolute values and binarized using proportional thresholding, retaining the top 10\% of strongest connections. Self-connections were removed before all network analyses.
	
	\subsubsection{Accuracy and Loss}
	We evaluated task performance primarily using accuracy and task loss. Accuracy was defined as the proportion of correctly classified trials. Loss was computed using categorical cross-entropy during both training and evaluation.
	
	Because different model variants used different regularization terms, the total loss values are not directly comparable.
	
	\subsubsection{Entropy}
	Entropy was used to quantify the degree of concentration versus randomness in the recurrent weight distribution (Fig.~\ref{fig:network-res}, Entropy). Lower entropy indicates a more concentrated and structured set of weights, whereas higher entropy indicates a more uniform, random-like distribution.
	
	To estimate this quantity, we computed the Shannon entropy of the recurrent weight distribution using Gaussian kernel density estimation. For each \(N \times N\) recurrent weight matrix,
	
	\begin{equation}
		H(W) = -\sum_i p(w_i)\log_2 p(w_i),
	\end{equation}
	
	where \(p(w_i)\) is the estimated probability density of the weights evaluated over 100 grid points spanning the observed range \cite{shannon_entropy_1948}.
	
	\subsubsection{Modularity}
	Modularity (\(Q\)) was used to quantify the extent to which the network organized into distinct communities (Fig.~\ref{fig:network-res}, Modularity). In neural systems, high modularity is often associated with functionally specialized subnetworks \cite{sporns_modular_2016}.
	
	We identified communities using the Clauset--Newman--Moore greedy modularity maximization algorithm \cite{clauset_modularity_2004}. This algorithm begins with each node assigned to its own community and iteratively merges the pair of communities that yields the largest increase in modularity until no further improvement is possible.
	
	Modularity was computed as
	
	\begin{equation}
		Q = \frac{1}{2m}\sum_{ij}\left[A_{ij} - \frac{k_i k_j}{2m}\right]\delta(c_i,c_j),
	\end{equation}
	
	where \(m\) is the total number of edges, \(A_{ij}\) is the adjacency matrix, \(k_i\) and \(k_j\) are the degrees of nodes \(i\) and \(j\), and \(\delta(c_i,c_j)\) equals 1 when nodes \(i\) and \(j\) belong to the same community and 0 otherwise \cite{newman_modularity_2006}.
	
	\subsubsection{Assortativity}
	Degree assortativity measures the extent to which nodes preferentially connect to other nodes of similar degree (Fig.~\ref{fig:network-res}, Assortativity). In our setting, it provides a compact description of whether the learned recurrent topology tends toward hub-rich integration or hub--periphery structure, distinguishing between segregation
	and integration.
	
	Assortativity was computed as \cite{newman_assortativity_2003}
	
	\begin{equation}
		r = \frac{\sum_{xy} xy \left( e_{xy} - a_x b_y \right)}{\sigma_a \sigma_b},
	\end{equation}
	
	where \(e_{xy}\) is the joint degree distribution over edges, \(a_x\) and \(b_y\) are the fractions of edges that originate from and terminate at nodes of degree \(x\) and \(y\), respectively, and \(\sigma_a\) and \(\sigma_b\) are the corresponding standard deviations.
	
	The coefficient satisfies \(-1 \le r \le 1\). Positive values ( $r > 0$) indicate \textit{assortative} organization, in which high-degree nodes connect preferentially to other high-degree nodes, forming clustered hubs that enhance global integration and efficient information transfer. Negative values ($r < 0$) indicate \textit{disassortative} organization, in which high-degree nodes connect primarily to low-degree nodes, producing hub-and-spoke topologies that promote modularity and specialization. Values near zero ($r \approx 0$) indicate the absence of strong degree-based preference (i.e. indicating presence of random connections).
	
	\subsubsection{Small-worldness}
	Small-worldness quantifies the extent to which a network combines strong local clustering with short global path lengths \cite{bassett_small-world_2006}, a property commonly observed in brain networks, where regions form tightly connected local modules while maintaining efficient long-range connections \cite{watts_collective_1998}. We quantified this property using  \cite{Humphries_smw_2008}:
	
	\begin{equation}
		\sigma = \frac{C / C_{\text{rand}}}{L / L_{\text{rand}}},
	\end{equation}
	
	where \(C\) is the mean clustering coefficient and \(L\) is the characteristic path length. The corresponding random-network baselines \(C_{\text{rand}}\) and \(L_{\text{rand}}\) were computed from 1,000 random binary graphs. A network was considered small-world when \(\sigma > 1\).
	
	The characteristic path length was defined as
	
	\begin{equation}
		L = \frac{1}{N(N-1)}\sum_{i \ne j} d_{ij},
	\end{equation}
	
	where \(N\) is the number of nodes and \(d_{ij}\) is the shortest path length between nodes \(i\) and \(j\).
	
	\subsection{Statistical Analysis}
	We analyzed mean metric values across model variants and tasks using a rank-based factorial ANOVA, implemented by ranking the dependent variable and then applying a standard three-way ANOVA to test the main and interaction effects of the three design factors \(W\), \(D\), and \(C\).
	
	To examine individual factor contributions in more detail, we then performed Kruskal--Wallis tests (\(H\)) \cite{kruskal_test_1952}, followed by pairwise Mann--Whitney post hoc comparisons (\(U\)) \cite{mann_whitney_1947}. Holm correction was applied to all pairwise comparisons \cite{holm_test_1979}. We report test statistics, \(p\)-values, and 95\% confidence intervals. For details, see the numerical values and corresponding tables in the supplementary material.
	
	\section*{Data Availability}
	Structural connectivity and functional activity data from the MICrONS dataset, along with links to APIs are available at \href{https://www.microns-explorer.org/cortical-mm3}{https://www.microns-explorer.org/cortical-mm3}. The calcium imaging data are available through the DANDI (Distributed Archive for Neurophysiology Data Integration) repository at \href{https://dandiarchive.org/dandiset/000402}{DANDISET~000402}. High‑resolution electron microscopy, segmentation, and morphological reconstructions of cortical circuits in mouse visual cortex are available through BossDB at \href{https://bossdb.org/project/microns-minnie}{https://bossdb.org/project/microns-minnie}. The processed data files generated from the functional computations used in the weight‑initialization procedure are available at \href{https://github.com/neurovium/CorticalBlueprintRNN}{\texttt{CorticalBlueprintRNN}}.
	
	\section*{Code Availability}
	The code used to preprocess the data, initialize model weights, and train the networks is available at \href{https://github.com/neurovium/CorticalBlueprintRNN}{\texttt{[github.com/neurovium/CorticalBlueprintRNN]}}. All scripts required to reproduce the analyses and results presented in this study are provided, along with documentation and example usage.
	
	
	
	\section*{Acknowledgments}
	N.D. wishes to acknowledge the support of NIH grant R24MH117295. M.S., R.R., and M.M. thank \href{https://neuromatch.io/}{Neuromatch Academy} for its support and resources for young scholars that was used for this study.
	
	\section*{References}
	
	\bibliography{CorticalBlueprintRNN}

@preamble{
 "\providecommand{\noopsort}[1]{}" 
 # "\providecommand{\singleletter}[1]{#1}%" 
}

@article{cutts_sttc_2014,
	title        = {Detecting Pairwise Correlations in Spike Trains: An Objective Comparison of Methods and Application to the Study of Retinal Waves},
	author       = {Cutts, Catherine S. and Eglen, Stephen J.},
	year         = 2014,
	journal      = {Journal of Neuroscience},
	publisher    = {Society for Neuroscience},
	volume       = 34,
	number       = 43,
	pages        = {14288--14303},
	doi          = {10.1523/JNEUROSCI.2767-14.2014},
	issn         = {0270-6474},
	url          = {https://doi.org/10.1523/JNEUROSCI.2767-14.2014}
}

@article{dorkenwald_cave_2025,
	title        = {CAVE: Connectome Annotation Versioning Engine},
	author       = {Dorkenwald, Sven and Schneider-Mizell, Casey M. and Brittain, Derrick and Halageri, Akhilesh and Jordan, Chris and Kemnitz, Nico and Castro, Manual A. and Silversmith, William and Maitin-Shephard, Jeremy and Troidl, Jakob and Pfister, Hanspeter and Gillet, Valentin and Xenes, Daniel and Bae, J. Alexander and Bodor, Agnes L. and Buchanan, JoAnn and Bumbarger, Daniel J. and Elabbady, Leila and Jia, Zhen and Kapner, Daniel and Kinn, Sam and Lee, Kisuk and Li, Kai and Lu, Ran and Macrina, Thomas and Mahalingam, Gayathri and Mitchell, Eric and Mondal, Shanka Subhra and Mu, Shang and Nehoran, Barak and Popovych, Sergiy and Takeno, Marc  and Torres, Russel and Turner, Nicholas L. and Wong, William and Wu, Jingpeng and Yin, Wenjing and Yu, Szi-chieh and Reid, R. Clay and da Costa, Nuno Ma{\c{c}}arico and Seung, H. Sebastian and Collman, Forrest},
	year         = 2025,
	journal      = {Nature Methods},
	volume       = 22,
	number       = 5,
	pages        = {1112--1120},
	doi          = {10.1038/s41592-024-02426-z},
	issn         = {1548-7105},
	url          = {https://doi.org/10.1038/s41592-024-02426-z}
}

@article{song_highly_2005,
	title        = {Highly Nonrandom Features of Synaptic Connectivity in Local Cortical Circuits},
	author       = {Song, Sen and Sj{\"o}str{\"o}m, Per Jesper and Reigl, Markus and Nelson, Sacha and Chklovskii, Dmitri B.},
	year         = 2005,
	journal      = {PLOS Biology},
	publisher    = {Public Library of Science},
	volume       = 3,
	number       = 3,
	pages        = {e68},
	doi          = {10.1371/journal.pbio.0030068},
	url          = {https://doi.org/10.1371/journal.pbio.0030068}
}

@article{loewenstein_multiplicative_2011,
	title        = {Multiplicative Dynamics Underlie the Emergence of the Log-Normal Distribution of Spine Sizes in the Neocortex In Vivo},
	author       = {Loewenstein, Yonatan and Kuras, Annerose and Rumpel, Simon},
	year         = 2011,
	journal      = {Journal of Neuroscience},
	publisher    = {Society for Neuroscience},
	volume       = 31,
	number       = 26,
	pages        = {9481--9488},
	doi          = {10.1523/JNEUROSCI.6130-10.2011},
	issn         = {0270-6474},
	url          = {https://doi.org/10.1523/JNEUROSCI.6130-10.2011}
}

@article{zhang_gonogo_2019,
	title        = {Active information maintenance in working memory by a sensory cortex},
	author       = {Zhang, Xiaoxing and Yan, Wenjun and Wang, Wenliang and Fan, Hongmei and Hou, Ruiqing and Chen, Yulei and Chen, Zhaoqin and Ge, Chaofan and Duan, Shumin and Compte, Albert and Li, Chengyu T.},
	year         = 2019,
	journal      = {eLife},
	publisher    = {eLife Sciences Publications, Ltd},
	volume       = 8,
	pages        = {e43191},
	doi          = {10.7554/eLife.43191},
	issn         = {2050-084X},
	url          = {https://doi.org/10.7554/eLife.43191}
}

@article{britten_pdm_1992,
	title        = {The analysis of visual motion: a comparison of neuronal and psychophysical performance},
	author       = {Britten, K. H. and Shadlen, M. N. and Newsome, W. T. and Movshon, J. A.},
	year         = 1992,
	journal      = {Journal of Neuroscience},
	publisher    = {Society for Neuroscience},
	volume       = 12,
	number       = 12,
	pages        = {4745--4765},
	doi          = {10.1523/JNEUROSCI.12-12-04745.1992},
	issn         = {0270-6474},
	url          = {https://doi.org/10.1523/JNEUROSCI.12-12-04745.1992}
}

@article{bassett_small-world_2006,
	title        = {Small-World Brain Networks},
	author       = {Bassett, Danielle Smith and Bullmore, Ed},
	year         = 2006,
	journal      = {The Neuroscientist},
	volume       = 12,
	number       = 6,
	pages        = {512--523},
	doi          = {10.1177/1073858406293182},
	url          = {https://doi.org/10.1177/1073858406293182}
}

@article{watts_collective_1998,
	title        = {Collective dynamics of `small-world' networks},
	author       = {Watts, Duncan J. and Strogatz, Steven H.},
	year         = 1998,
	journal      = {Nature},
	volume       = 393,
	number       = 6684,
	pages        = {440--442},
	doi          = {10.1038/30918},
	issn         = {1476-4687},
	url          = {https://doi.org/10.1038/30918}
}

@article{clauset_modularity_2004,
	title        = {Finding community structure in very large networks},
	author       = {Clauset, Aaron and Newman, M. E. J. and Moore, Cristopher},
	year         = 2004,
	journal      = {Physical Review E},
	publisher    = {American Physical Society},
	volume       = 70,
	number       = 6,
	pages        = {066111},
	doi          = {10.1103/PhysRevE.70.066111},
	url          = {https://doi.org/10.1103/PhysRevE.70.066111}
}

@article{sporns_modular_2016,
	title        = {Modular Brain Networks},
	author       = {Sporns, Olaf and Betzel, Richard F.},
	year         = 2016,
	journal      = {Annual Review of Psychology},
	volume       = 67,
	number       = 1,
	pages        = {613--640},
	doi          = {10.1146/annurev-psych-122414-033634},
	issn         = {0066-4308},
	url          = {https://doi.org/10.1146/annurev-psych-122414-033634}
}

@article{Pearson_Corr_1895,
	title        = {VII. Note on regression and inheritance in the case of two parents},
	author       = {Pearson, Karl},
	year         = 1895,
	journal      = {Proceedings of the Royal Society of London},
	volume       = 58,
	number       = {347-352},
	pages        = {240--242},
	doi          = {10.1098/rspl.1895.0041},
	issn         = {0370-1662},
	url          = {https://doi.org/10.1098/rspl.1895.0041}
}

@article{Humphries_smw_2008,
	title        = {Network `Small-World-Ness': A Quantitative Method for Determining Canonical Network Equivalence},
	author       = {Humphries, Mark D. and Gurney, Kevin},
	year         = 2008,
	journal      = {PLOS ONE},
	publisher    = {Public Library of Science},
	volume       = 3,
	number       = 4,
	pages        = {e2051},
	doi          = {10.1371/journal.pone.0002051},
	url          = {https://doi.org/10.1371/journal.pone.0002051}
}

@article{newman_modularity_2006,
	title        = {Modularity and community structure in networks},
	author       = {Newman, M. E. J.},
	year         = 2006,
	journal      = {Proceedings of the National Academy of Sciences},
	volume       = 103,
	number       = 23,
	pages        = {8577--8582},
	doi          = {10.1073/pnas.0601602103},
	url          = {https://doi.org/10.1073/pnas.0601602103}
}

@article{shannon_entropy_1948,
	title        = {A Mathematical Theory of Communication},
	author       = {Shannon, C. E.},
	year         = 1948,
	journal      = {Bell System Technical Journal},
	volume       = 27,
	number       = 3,
	pages        = {379--423},
	doi          = {10.1002/j.1538-7305.1948.tb01338.x},
	url          = {https://doi.org/10.1002/j.1538-7305.1948.tb01338.x}
}

@article{newman_assortativity_2003,
	title        = {Mixing patterns in networks},
	author       = {Newman, M. E. J.},
	year         = 2003,
	journal      = {Physical Review E},
	publisher    = {American Physical Society},
	volume       = 67,
	number       = 2,
	pages        = {026126},
	doi          = {10.1103/PhysRevE.67.026126},
	url          = {https://doi.org/10.1103/PhysRevE.67.026126}
}

@book{dayan_comp_2005,
	title        = {Theoretical Neuroscience: Computational and Mathematical Modeling of Neural Systems},
	author       = {Dayan, Peter and Abbott, L. F.},
	year         = 2005,
	publisher    = {The MIT Press},
	address      = {Cambridge, MA},
	isbn         = {0262541858}
}

@article{richards_deep_2019,
	title        = {A deep learning framework for neuroscience},
	author       = {Richards, Blake A. and Lillicrap, Timothy P. and Beaudoin, Philippe and Bengio, Yoshua and Bogacz, Rafal and Christensen, Amelia and Clopath, Claudia and Costa, Rui Ponte and de Berker, Archy and Ganguli, Surya and Gillon, Colleen J. and Hafner, Danijar and Kepecs, Adam and Kriegeskorte, Nikolaus and Latham, Peter and Lindsay, Grace W. and Miller, Kenneth D. and Naud, Richard and Pack, Christopher C. and Poirazi, Panayiota and Roelfsema, Pieter and Sacramento, Jo{\~a}o and Saxe, Andrew and Scellier, Benjamin and Schapiro, Anna C. and Senn, Walter and Wayne, Greg and Yamins, Daniel and Zenke, Friedemann and Zylberberg, Joel and Therien, Denis and Kording, Konrad P.},
	year         = 2019,
	journal      = {Nature Neuroscience},
	publisher    = {Nature Publishing Group},
	volume       = 22,
	number       = 11,
	pages        = {1761--1770},
	doi          = {10.1038/s41593-019-0520-2},
	url          = {https://doi.org/10.1038/s41593-019-0520-2}
}

@article{kruskal_test_1952,
	title        = {Use of Ranks in One-Criterion Variance Analysis},
	author       = {Kruskal, William H. and Wallis, W. Allen},
	year         = 1952,
	journal      = {Journal of the American Statistical Association},
	publisher    = {American Statistical Association},
	volume       = 47,
	number       = 260,
	pages        = {583--621},
	doi          = {10.1080/01621459.1952.10483441},
	url          = {https://doi.org/10.1080/01621459.1952.10483441}
}

@article{mann_whitney_1947,
	title        = {On a Test of Whether one of Two Random Variables is Stochastically Larger than the Other},
	author       = {Mann, H. B. and Whitney, D. R.},
	year         = 1947,
	journal      = {The Annals of Mathematical Statistics},
	publisher    = {Institute of Mathematical Statistics},
	volume       = 18,
	number       = 1,
	pages        = {50--60},
	doi          = {10.1214/aoms/1177730491},
	url          = {https://doi.org/10.1214/aoms/1177730491}
}

@article{holm_test_1979,
	title        = {A Simple Sequentially Rejective Multiple Test Procedure},
	author       = {Holm, Sture},
	year         = 1979,
	journal      = {Scandinavian Journal of Statistics},
	publisher    = {[Board of the Foundation of the Scandinavian Journal of Statistics, Wiley]},
	volume       = 6,
	number       = 2,
	pages        = {65--70},
	url          = {http://www.jstor.org/stable/4615733}
}

@article{bullmore2012theeconomyof,
	title        = {The economy of brain network organization},
	author       = {Bullmore, Ed and Sporns, Olaf},
	year         = 2012,
	journal      = {Nature Reviews Neuroscience},
	publisher    = {Springer Nature},
	volume       = 13,
	number       = 5,
	pages        = {336--349},
	doi          = {10.1038/nrn3214},
	issn         = {1471-003X},
	url          = {https://doi.org/10.1038/nrn3214}
}

@article{samu2014influenceofwiring,
	title        = {Influence of Wiring Cost on the Large-Scale Architecture of Human Cortical Connectivity},
	author       = {Samu, David and Seth, Anil K. and Nowotny, Thomas},
	year         = 2014,
	journal      = {PLOS Computational Biology},
	publisher    = {Public Library of Science},
	volume       = 10,
	number       = 4,
	pages        = {e1003557},
	doi          = {10.1371/journal.pcbi.1003557},
	issn         = {1553-7358},
	url          = {https://doi.org/10.1371/journal.pcbi.1003557}
}

@article{budd2012communicationandwiring,
	title        = {Communication and wiring in the cortical connectome},
	author       = {Budd, Julian M. L. and Kisv{\'a}rday, Zolt{\'a}n F.},
	year         = 2012,
	journal      = {Frontiers in Neuroanatomy},
	publisher    = {Frontiers Media SA},
	volume       = 6,
	pages        = 42,
	doi          = {10.3389/fnana.2012.00042},
	issn         = {1662-5129},
	url          = {https://doi.org/10.3389/fnana.2012.00042}
}

@article{schroter2017microconnectomicsprobingthe,
	title        = {Micro-connectomics: probing the organization of neuronal networks at the cellular scale},
	author       = {Schr{\"o}ter, Manuel and Paulsen, Ole and Bullmore, Edward T.},
	year         = 2017,
	journal      = {Nature Reviews Neuroscience},
	publisher    = {Springer Nature},
	volume       = 18,
	number       = 3,
	pages        = {131--146},
	doi          = {10.1038/nrn.2016.182},
	issn         = {1471-003X},
	url          = {https://doi.org/10.1038/nrn.2016.182}
}

@article{shipp2007structureandfunction,
	title        = {Structure and function of the cerebral cortex},
	author       = {Shipp, Stewart},
	year         = 2007,
	journal      = {Current Biology},
	publisher    = {Elsevier BV},
	volume       = 17,
	number       = 12,
	pages        = {R443--R449},
	doi          = {10.1016/j.cub.2007.03.044},
	issn         = {0960-9822},
	url          = {https://doi.org/10.1016/j.cub.2007.03.044}
}

@article{achterberg2023spatiallyembeddedrecurrent,
	title        = {Spatially embedded recurrent neural networks reveal widespread links between structural and functional neuroscience findings},
	author       = {Achterberg, Jascha and Akarca, Danyal and Strouse, D. J. and Duncan, John and Astle, Duncan E.},
	year         = 2023,
	journal      = {Nature Machine Intelligence},
	publisher    = {Springer Nature},
	volume       = 5,
	number       = 12,
	pages        = {1369--1381},
	doi          = {10.1038/s42256-023-00748-9},
	issn         = {2522-5839},
	url          = {https://doi.org/10.1038/s42256-023-00748-9}
}

@article{khona2023winningthelottery,
	title        = {Winning the Lottery With Neural Connectivity Constraints: Faster Learning Across Cognitive Tasks With Spatially Constrained Sparse RNNs},
	author       = {Khona, Mikail and Chandra, Sarthak and Ma, Joy J. and Fiete, Ila R.},
	year         = 2023,
	journal      = {Neural Computation},
	publisher    = {MIT Press},
	volume       = 35,
	number       = 11,
	pages        = {1850--1869},
	doi          = {10.1162/neco_a_01613},
	issn         = {0899-7667},
	url          = {https://doi.org/10.1162/neco_a_01613}
}

@article{barak2017recurrentneuralnetworks,
	title        = {Recurrent neural networks as versatile tools of neuroscience research},
	author       = {Barak, Omri},
	year         = 2017,
	journal      = {Current Opinion in Neurobiology},
	publisher    = {Elsevier BV},
	volume       = 46,
	pages        = {1--6},
	doi          = {10.1016/j.conb.2017.06.003},
	issn         = {0959-4388},
	url          = {https://doi.org/10.1016/j.conb.2017.06.003}
}

@misc{schuessler2020theinterplaybetween,
	title        = {The interplay between randomness and structure during learning in RNNs},
	author       = {Schuessler, Friedrich and Mastrogiuseppe, Francesca and Dubreuil, Alexis and Ostojic, Srdjan and Barak, Omri},
	year         = 2021,
	url          = {https://arxiv.org/abs/2006.11036},
	eprint       = {2006.11036},
	archiveprefix = {arXiv},
	primaryclass = {q-bio.NC}
}

@article{krause2022operativedimensionsin,
	title        = {Operative dimensions in unconstrained connectivity of recurrent neural networks},
	author       = {Krause, Renate and Cook, Matthew and Kollmorgen, Sepp and Mante, Valerio and Indiveri, Giacomo},
	year         = 2022,
	journal      = {bioRxiv},
	publisher    = {Cold Spring Harbor Laboratory},
	doi          = {10.1101/2022.06.03.494670},
	url          = {https://doi.org/10.1101/2022.06.03.494670}
}

@article{beiran2025predictionofneural,
	title        = {Prediction of neural activity in connectome-constrained recurrent networks},
	author       = {Beiran, Manuel and Litwin-Kumar, Ashok},
	year         = 2025,
	journal      = {Nature Neuroscience},
	volume       = 28,
	number       = 12,
	pages        = {2561--2574},
	doi          = {10.1038/s41593-025-02080-4},
	issn         = {1546-1726},
	url          = {https://doi.org/10.1038/s41593-025-02080-4}
}

@article{pulvermuller2021biologicalconstraintson,
	title        = {Biological constraints on neural network models of cognitive function},
	author       = {Pulverm{\"u}ller, Friedemann and Tomasello, Rosario and Henningsen-Schomers, Malte R. and Wennekers, Thomas},
	year         = 2021,
	journal      = {Nature Reviews Neuroscience},
	publisher    = {Springer Nature},
	volume       = 22,
	number       = 8,
	pages        = {488--502},
	doi          = {10.1038/s41583-021-00473-5},
	issn         = {1471-003X},
	url          = {https://doi.org/10.1038/s41583-021-00473-5}
}

@article{marblestone2016towardanintegration,
	title        = {Toward an Integration of Deep Learning and Neuroscience},
	author       = {Marblestone, Adam H. and Wayne, Greg and Kording, Konrad P.},
	year         = 2016,
	journal      = {Frontiers in Computational Neuroscience},
	publisher    = {Frontiers Media SA},
	volume       = 10,
	pages        = 94,
	doi          = {10.3389/fncom.2016.00094},
	issn         = {1662-5188},
	url          = {https://doi.org/10.3389/fncom.2016.00094}
}

@article{bae2025functionalconnectomicsspanning,
	title        = {Functional connectomics spanning multiple areas of mouse visual cortex},
	author       = {Bae, J. Alexander and Baptiste, Mahaly and Bodor, Agnes L. and Brittain, Derrick and Buchanan, JoAnn and Bumbarger, Daniel J. and Castro, Manuel A. and Celii, Brendan and Cobos, Erick and Collman, Forrest and da Costa, Nuno Ma{\c{c}}arico and Dorkenwald, Sven and Elabbady, Leila and Fahey, Paul G. and Fliss, Tim and Froudakis, Emmanouil and Gager, Jay and Gamlin, Clare and Halageri, Akhilesh and Hebditch, James and Jia, Zhen  and Jordan, Chris and Kapner, Daniel and Kemnitz, Nico and Kinn, Sam and Koolman, Selden and Kuehner, Kai and Lee, Kisuk  and Li, Kai and Lu, Ran  and Macrina, Thomas and Mahalingam, Gayathri and McReynolds, Sarah and Miranda, Elanine and Mitchell, Eric  and Mondal, Shanka Subhra and Moore, Merlin and Mu, Shang and Muhammad, Taliah and Nehoran, Barak and Ogedengbe, Oluwaseun and Papadopoulos, Christos and Papadopoulos, Stelios and Patel, Saumil S. and Pitkow, Xaq and Popovych, Sergiy and Ramos, Anthony and Reid, R. Clay and Reimer, Jacob and Schneider-Mizell, Casey and Seung, H. Sebastian and Silverman, Ben and Silversmith, William and Sterling, Amy and Sinz, Fabian H. and Smith, Cameron L. and Suckow, Shelby  and Tan, Z. H. and Tolias, Andreas S. and Torres, Russel and Turner, Nicholas L. and Walker, Edgar Y. and Wang, Tianyu and Williams, G. and Williams, S. and Willie, K. and Willie, Ryan and Wong, William and Wu, Jingpeng and Xu, Chris and Yang, Runzhe and Yatsenko, Dimitri and Ye, Fei and Yin, Wenjing and Yu, Szi-chieh},
	year         = 2025,
	journal      = {Nature},
	volume       = 640,
	pages        = {435--447},
	doi          = {10.1038/s41586-025-08790-w},
	url          = {https://doi.org/10.1038/s41586-025-08790-w}
}

@article{turner2020multiscaleandmultimodal,
	title        = {Multiscale and multimodal reconstruction of cortical structure and function},
	author       = {Turner, Nicholas L. and Macrina, Thomas and Bae, J. Alexander and Yang, Runzhe and Wilson, Alyssa M. and Schneider-Mizell, Casey and Lee, Kisuk and Lu, Ran and Wu, Jingpeng and Bodor, Agnes L. and Bleckert, Adam A. and Brittain, Derrick and Froudarakis, Emmanouil and Dorkenwald, Sven and Collman, Forrest and Kemnitz, Nico and Ih, Dodam and Silversmith, William M. and Zung, Jonathan and Zlateski, Aleksandar and Tartavull, Ignacio and Yu, Szi-chieh and Popovych, Sergiy and Mu, Shang and Wong, William and Jordan, Chris S. and Castro, Manuel and Buchanan, JoAnn and Bumbarger, Daniel J. and Takeno, Marc  and Torres, Russel and Mahalingam, Gayathri and Elabbady, Leila and Li, Yang and Cobos, Erick and Zhou, Pengcheng and Suckow, Shelby and Becker, Lynne and Paninski, Liam and Polleux, Franck and Reimer, Jacob  and Tolias, Andreas S. and Reid, R. Clay and da Costa, Nuno Ma{\c{c}}arico and Seung, H. Sebastian},
	year         = 2020,
	journal      = {bioRxiv},
	publisher    = {Cold Spring Harbor Laboratory},
	doi          = {10.1101/2020.10.14.338681},
	url          = {https://doi.org/10.1101/2020.10.14.338681}
}

@article{lappalainen2024connectomeconstrainednetworkspredict,
	title        = {Connectome-constrained networks predict neural activity across the fly visual system},
	author       = {Lappalainen, Janne K. and Tschopp, Fabian D. and Prakhya, Sridhama and McGill, Mason and Nern, Aljoscha and Shinomiya, Kazunori and Takemura, Shin-ya  and Gruntman, Eyal and Macke, Jakob H. and Turaga, Srinivas C.},
	year         = 2024,
	journal      = {Nature},
	publisher    = {Springer Nature},
	volume       = 634,
	number       = 8036,
	pages        = {1132--1140},
	doi          = {10.1038/s41586-024-07939-3},
	issn         = {0028-0836},
	url          = {https://doi.org/10.1038/s41586-024-07939-3}
}

@article{wang2016brainstructureand,
	title        = {Brain structure and dynamics across scales: in search of rules},
	author       = {Wang, Xiao-Jing and Kennedy, Henry},
	year         = 2016,
	journal      = {Current Opinion in Neurobiology},
	publisher    = {Elsevier BV},
	volume       = 37,
	pages        = {92--98},
	doi          = {10.1016/j.conb.2015.12.010},
	issn         = {0959-4388},
	url          = {https://doi.org/10.1016/j.conb.2015.12.010}
}

@article{Liu2025benchmarkconnectivity,
	title        = {Benchmarking methods for mapping functional connectivity in the brain},
	author       = {Liu, Zhen-Qi and Luppi, Andrea I. and Hansen, Justine Y. and Tian, Ye Ella and Zalesky, Andrew and Yeo, B. T. Thomas and Fulcher, Ben D. and Misi{\'c}, Bratislav},
	year         = 2025,
	journal      = {Nature Methods},
	volume       = 22,
	number       = 7,
	pages        = {1593--1602},
	doi          = {10.1038/s41592-025-02704-4},
	issn         = {1548-7105},
	url          = {https://doi.org/10.1038/s41592-025-02704-4}
}

@misc{sheeran2024spatialembeddingpromotes,
	title        = {Spatial embedding promotes a specific form of modularity with low entropy and heterogeneous spectral dynamics},
	author       = {Sheeran, Cornelia and Ham, Andrew S. and Astle, Duncan E. and Achterberg, Jascha and Akarca, Danyal},
	year         = 2024,
	url          = {https://arxiv.org/abs/2409.17693},
	eprint       = {2409.17693},
	archiveprefix = {arXiv},
	primaryclass = {cs.NE}
}

@article{song2016trainingexcitatoryinhibitoryrecurrent,
	title        = {Training Excitatory-Inhibitory Recurrent Neural Networks for Cognitive Tasks: A Simple and Flexible Framework},
	author       = {Song, H. F. and Yang, G. R. and Wang, Xiao-Jing},
	year         = 2016,
	journal      = {PLOS Computational Biology},
	publisher    = {Public Library of Science},
	volume       = 12,
	number       = 2,
	pages        = {e1004792},
	doi          = {10.1371/journal.pcbi.1004792},
	url          = {https://doi.org/10.1371/journal.pcbi.1004792}
}

@inproceedings{li2023learningbetterwith,
	title        = {Learning better with Dale{\textquoteright}s Law: A Spectral Perspective},
	author       = {Li, Pingsheng and Cornford, Jonathan and Ghosh, Arna and Richards, Blake},
	year         = 2023,
	booktitle    = {Thirty-seventh Conference on Neural Information Processing Systems},
	url          = {https://openreview.net/forum?id=rDiMgZulwi}
}

@article{balwani2025constructingbiologicallyconstrained,
	title        = {Constructing biologically constrained RNNs via Dale’s backpropagation and topologically informed pruning},
	author       = {Balwani, Aishwarya and Wang, Alex Q. and Najafi, Farzaneh and Choi, Hannah},
	year         = 2025,
	journal      = {Science Advances},
	volume       = 11,
	number       = 50,
	pages        = {eadw4970},
	doi          = {10.1126/sciadv.adw4970},
	url          = {https://doi.org/10.1126/sciadv.adw4970}
}

@article{meunier2010modularandhierarchically,
	title        = {Modular and Hierarchically Modular Organization of Brain Networks},
	author       = {Meunier, David  and Lambiotte, Renaud and Bullmore, Edward T.},
	year         = 2010,
	journal      = {Frontiers in Neuroscience},
	publisher    = {Frontiers Media SA},
	volume       = 4,
	pages        = 200,
	doi          = {10.3389/fnins.2010.00200},
	issn         = {1662-4548},
	url          = {https://doi.org/10.3389/fnins.2010.00200}
}

@inproceedings{rovny2024connectomeconstrainedspatiallyembedded,
	title        = {Connectome-constrained spatially embedded recurrent neural networks},
	author       = {Rovn{\'y}, Maro{\v{s}} and Akarca, Danyal and Achterberg, Jascha and Astle, Duncan},
	year         = 2024,
	booktitle    = {Proceedings of the Computational and Cognitive Neuroscience (CCN) Meeting},
	address      = {Boston, MA, USA},
	url          = {https://2024.ccneuro.org/pdf/618_Paper_authored_CCN24.pdf}
}

@article{park2013structuralandfunctional,
	title        = {Structural and Functional Brain Networks: From Connections to Cognition},
	author       = {Park, Hae-Jeong and Friston, Karl J.},
	year         = 2013,
	journal      = {Science},
	publisher    = {American Association for the Advancement of Science},
	volume       = 342,
	number       = 6158,
	pages        = 1238411,
	doi          = {10.1126/science.1238411},
	url          = {https://doi.org/10.1126/science.1238411}
}

@article{lynn2019thephysicsof,
	title        = {The physics of brain network structure, function and control},
	author       = {Lynn, Christopher W. and Bassett, Danielle S.},
	year         = 2019,
	journal      = {Nature Reviews Physics},
	publisher    = {Springer Nature},
	volume       = 1,
	number       = 5,
	pages        = {318--332},
	doi          = {10.1038/s42254-019-0040-8},
	url          = {https://doi.org/10.1038/s42254-019-0040-8}
}

@article{Yanez2026morphoelectricInhibitory,
	title        = {Morphoelectric properties of inhibitory neurons shift gradually and regardless of cell type along the depth of the cerebral cortex},
	author       = {Y{\'a}{\~n}ez, Felipe and Messore, Fernando and Qi, Guanxiao and Dehghani, Nima and Meyer, Hanno S. and Feldmeyer, Dirk and Sakmann, Bert and Oberlaender, Marcel},
	year         = 2026,
	journal      = {bioRxiv},
	publisher    = {Cold Spring Harbor Laboratory},
	doi          = {10.64898/2026.03.05.709819},
	url          = {https://doi.org/10.64898/2026.03.05.709819}
}

@article{schneidermizell2024celltypespecificinhibitorycircuitry,
	title        = {Inhibitory specificity from a connectomic census of mouse visual cortex},
	author       = {Schneider-Mizell, Casey M. and Bodor, Agnes L. and Brittain, Derrick and Buchanan, JoAnn and Bumbarger, Daniel J. and Elabbady, Leila and Gamlin, Clare and Kapner, Daniel and Kinn, Sam and Mahalingam, Gayathri and Seshamani, Sharmishtaa  and Suckow, Shelby and Takeno, Marc and Torres, Russel and Yin, Wenjing and Dorkenwald, Sven and Bae, J. Alexander and Castro, Manuel A. and Halageri, Akhilesh and Jia, Zhen and Jordan, Chris and Kemnitz, Nico and Lee, Kisuk and Li, Kai and Lu, Ran and Macrina, Thomas and Mitchell, Eric and Mondal, Shanka Subhra and Mu, Shang and Nehoran, Barak and Popovych, Sergiy and Silversmith, William and Turner, Nicholas L. and Wong, William and Wu, Jingpeng and Reimer, Jacob  and Tolias, Andreas S. and Seung, H. Sebastian and Reid, R. Clay and Collman, Forrest and da Costa, Nuno Ma{\c{c}}arico},
	year         = 2025,
	journal      = {Nature},
	volume       = 640,
	number       = 8058,
	pages        = {448--458},
	doi          = {10.1038/s41586-024-07780-8},
	issn         = {1476-4687},
	url          = {https://doi.org/10.1038/s41586-024-07780-8}
}

@article{petersen2015brainnetworksand,
	title        = {Brain Networks and Cognitive Architectures},
	author       = {Petersen, Steven E. and Sporns, Olaf},
	year         = 2015,
	journal      = {Neuron},
	publisher    = {Elsevier BV},
	volume       = 88,
	number       = 1,
	pages        = {207--219},
	doi          = {10.1016/j.neuron.2015.09.027},
	issn         = {0896-6273},
	url          = {https://doi.org/10.1016/j.neuron.2015.09.027}
}

@article{Matelsky2021motif,
	title        = {DotMotif: an open-source tool for connectome subgraph isomorphism search and graph queries},
	author       = {Matelsky, Jordan K. and Reilly, Elizabeth P. and Johnson, Erik C. and Stiso, Jennifer and Bassett, Danielle S. and Wester, Brock A. and Gray-Roncal, William},
	year         = 2021,
	journal      = {Scientific Reports},
	volume       = 11,
	number       = 1,
	pages        = 13045,
	doi          = {10.1038/s41598-021-91025-5},
	issn         = {2045-2322},
	url          = {https://doi.org/10.1038/s41598-021-91025-5}
}

@article{Vanderhaeghen2023evo,
	title        = {Developmental mechanisms underlying the evolution of human cortical circuits},
	author       = {Vanderhaeghen, Pierre and Polleux, Franck},
	year         = 2023,
	journal      = {Nature Reviews Neuroscience},
	volume       = 24,
	number       = 4,
	pages        = {213--232},
	doi          = {10.1038/s41583-023-00675-z},
	issn         = {1471-0048},
	url          = {https://doi.org/10.1038/s41583-023-00675-z}
}

@article{Chklovskii2002wiringOptim,
	title        = {Wiring Optimization in Cortical Circuits},
	author       = {Chklovskii, Dmitri B. and Schikorski, Thomas and Stevens, Charles F.},
	year         = 2002,
	journal      = {Neuron},
	volume       = 34,
	number       = 3,
	pages        = {341--347},
	doi          = {10.1016/S0896-6273(02)00679-7},
	issn         = {0896-6273},
	url          = {https://doi.org/10.1016/S0896-6273(02)00679-7}
}

@article{Ding2025WiringRule,
	title        = {Functional connectomics reveals general wiring rule in mouse visual cortex},
	author       = {Ding, Zhuokun and Fahey, Paul G. and Papadopoulos, Stelios and Wang, Eric Y. and Celii, Brendan and Papadopoulos, Christos and Chang, Andersen and Kunin, Alexander B. and Tran, Dat and Fu, Jiakun and Ding, Zhiwei and Patel, Saumil and Ntanavara, Lydia and Froebe, Rachel and Ponder, Kayla and Muhammad, Taliah and Bae, J. Alexander and Bodor, Agnes L. and Brittain, Derrick and Buchanan, JoAnn and Bumbarger, Daniel J. and Castro, Manuel A. and Cobos, Erick  and Dorkenwald, Sven and Elabbady, Leila and Halageri, Akhilesh and Jia, Zhen and Jordan, Chris and Kapner, Dan and Kemnitz, Nico and Kinn, Sam and Lee, Kisuk and Li, Kai and Lu, Ran and Macrina, Thomas and Mahalingam, Gayathri and Mitchell, Eric and Mondal, Shanka Subhra and Mu, Shang and Nehoran, Barak and Popovych, Sergiy and Schneider-Mizell, Casey M. and Silversmith, William and Takeno, Marc and Torres, Russel and Turner, Nicholas L. and Wong, William and Wu, Jingpeng and Yin, Wenjing and Yu, Szi-chieh  and Yatsenko, Dimitri and Froudarakis, Emmanouil and Sinz, Fabian and Josi{\'{c}}, Kre{\v{s}}imir and Rosenbaum, Robert and Seung, H. Sebastian and Collman, Forrest and da Costa, Nuno Ma{\c{c}}arico and Reid, R. Clay and Walker, Edgar Y. and Pitkow, Xaq and Reimer, Jacob  and Tolias, Andreas S.},
	year         = 2025,
	journal      = {Nature},
	volume       = 640,
	number       = 8058,
	pages        = {459--469},
	doi          = {10.1038/s41586-025-08840-3},
	issn         = {1476-4687},
	url          = {https://doi.org/10.1038/s41586-025-08840-3}
}

@article{Dorkenwald2024connectomeWire,
	title        = {Neuronal wiring diagram of an adult brain},
	author       = {Dorkenwald, Sven and Matsliah, Arie and Sterling, Amy R. and Schlegel, Philipp and Yu, Szi-chieh and McKellar, Claire E. and Lin, Albert and Costa, Marta and Eichler, Katharina and Yin, Yijie and Silversmith, Will and Schneider-Mizell, Casey and Jordan, Chris S. and Brittain, Derrick and Halageri, Akhilesh and Kuehner, Kai and Ogedengbe, Oluwaseun and Morey, Ryan and Gager, Jay and Kruk, Krzysztof and Perlman, Eric and Yang, Runzhe and Deutsch, David and Bland, Doug and Sorek, Marissa and Lu, Ran and Macrina, Thomas and Lee, Kisuk and Bae, J. Alexander and Mu, Shang and Nehoran, Barak and Mitchell, Eric and Popovych, Sergiy and Wu, Jingpeng and Jia, Zhen and Castro, Manuel A. and Kemnitz, Nico and Ih, Dodam and Bates, Alexander Shakeel and Eckstein, Nils  and Funke, Jan and Collman, Forrest and Bock, Davi D. and Jefferis, Gregory S. X. E. and Seung, H. Sebastian and Murthy, Mala and Consortium, The FlyWire},
	year         = 2024,
	journal      = {Nature},
	volume       = 634,
	number       = 8032,
	pages        = {124--138},
	doi          = {10.1038/s41586-024-07558-y},
	issn         = {1476-4687},
	url          = {https://doi.org/10.1038/s41586-024-07558-y}
}

@article{Cuntz2010GrowingWire,
	title        = {One Rule to Grow Them All: A General Theory of Neuronal Branching and Its Practical Application},
	author       = {Cuntz, Hermann and Forstner, Friedrich and Borst, Alexander and H{\"a}usser, Michael},
	year         = 2010,
	journal      = {PLOS Computational Biology},
	publisher    = {Public Library of Science},
	volume       = 6,
	number       = 8,
	pages        = {e1000877},
	doi          = {10.1371/journal.pcbi.1000877},
	url          = {https://doi.org/10.1371/journal.pcbi.1000877}
}

@article{Aprile2022smallworldOptim,
	title        = {The small world coefficient optimizes information processing in 2D neuronal networks},
	author       = {Aprile, F. and Onesto, V. and Gentile, F.},
	year         = 2022,
	journal      = {npj Systems Biology and Applications},
	volume       = 8,
	number       = 1,
	pages        = 4,
	doi          = {10.1038/s41540-022-00215-y},
	issn         = {2056-7189},
	url          = {https://doi.org/10.1038/s41540-022-00215-y}
}

@article{Latora2001smallworldEfficient,
	title        = {Efficient Behavior of Small-World Networks},
	author       = {Latora, Vito and Marchiori, Massimo},
	year         = 2001,
	journal      = {Physical Review Letters},
	publisher    = {American Physical Society},
	volume       = 87,
	number       = 19,
	pages        = 198701,
	doi          = {10.1103/PhysRevLett.87.198701},
	url          = {https://doi.org/10.1103/PhysRevLett.87.198701}
}

@article{Newsome1988,
	title        = {A selective impairment of motion perception following lesions of the middle temporal visual area (MT)},
	author       = {Newsome, William T. and Par{\'e}, Edmond B.},
	year         = 1988,
	journal      = {The Journal of Neuroscience},
	volume       = 8,
	number       = 6,
	pages        = {2201--2211},
	doi          = {10.1523/JNEUROSCI.08-06-02201.1988},
	url          = {https://doi.org/10.1523/JNEUROSCI.08-06-02201.1988}
}

@article{Shadlen2001,
	title        = {Neural Basis of a Perceptual Decision in the Parietal Cortex (Area LIP) of the Rhesus Monkey},
	author       = {Shadlen, Michael N. and Newsome, William T.},
	year         = 2001,
	journal      = {Journal of Neurophysiology},
	volume       = 86,
	number       = 4,
	pages        = {1916--1936},
	doi          = {10.1152/jn.2001.86.4.1916},
	url          = {https://doi.org/10.1152/jn.2001.86.4.1916}
}

@misc{johnson2023exploitinglargeneuroimagingdatasets,
	title        = {Exploiting Large Neuroimaging Datasets to Create Connectome-Constrained Approaches for more Robust, Efficient, and Adaptable Artificial Intelligence},
	author       = {Johnson, Erik C. and Robinson, Brian S. and Vallabha, Gautam K. and Joyce, Justin and Matelsky, Jordan K. and Norman-Tenazas, Raphael and Western, Isaac and Villafa{\~n}e-Delgado, Marisel and Cervantes, Martha and Robinette, Michael S. and Reddy, Arun V. and Kitchell, Lindsey and Rivlin, Patricia K. and Reilly, Elizabeth P. and Drenkow, Nathan and Roos, Matthew J. and Wang, I-Jeng and Wester, Brock A. and Gray-Roncal, William R. and Hoffmann, Joan A.},
	year         = 2023,
	url          = {https://arxiv.org/abs/2305.17300},
	eprint       = {2305.17300},
	archiveprefix = {arXiv},
	primaryclass = {cs.NE}
}

@article{Clune2013evomodularity,
	title        = {The evolutionary origins of modularity},
	author       = {Clune, Jeff and Mouret, Jean-Baptiste and Lipson, Hod},
	year         = 2013,
	journal      = {Proceedings of the Royal Society B: Biological Sciences},
	volume       = 280,
	number       = 1755,
	pages        = 20122863,
	doi          = {10.1098/rspb.2012.2863},
	issn         = {0962-8452},
	url          = {https://doi.org/10.1098/rspb.2012.2863}
}

@article{Zhang2025wiringeconomy,
	title        = {Brain-inspired wiring economics for artificial neural networks},
	author       = {Zhang, Xin-Jie and Moore, Jack Murdoch and Gao, Ting-Ting and Zhang, Xiaozhu and Yan, Gang},
	year         = 2025,
	journal      = {PNAS Nexus},
	volume       = 4,
	number       = 1,
	pages        = {pgae580},
	doi          = {10.1093/pnasnexus/pgae580},
	issn         = {2752-6542},
	url          = {https://doi.org/10.1093/pnasnexus/pgae580}
}

@article{Dawid1979conditionalindepence,
    author = {Dawid, A. P.},
    title = {Conditional Independence in Statistical Theory},
    journal = {Journal of the Royal Statistical Society: Series B (Methodological)},
    volume = {41},
    number = {1},
    pages = {1-15},
    year = {1979},
    month = {09},
    issn = {0035-9246},
    doi = {10.1111/j.2517-6161.1979.tb01052.x},
    url = {https://doi.org/10.1111/j.2517-6161.1979.tb01052.x},
}

@article{Liegeois2020funccon,
    author = {Liégeois, Raphael and Santos, Augusto and Matta, Vincenzo and Van De Ville, Dimitri and Sayed, Ali H.},
    title = {Revisiting correlation-based functional connectivity and its relationship with structural connectivity},
    journal = {Network Neuroscience},
    volume = {4},
    number = {4},
    pages = {1235-1251},
    year = {2020},
    month = {12},
    issn = {2472-1751},
    doi = {10.1162/netn_a_00166},
    url = {https://doi.org/10.1162/netn_a_00166},
}

@article{Das2017precisionmatrix,
    author = {Das, Anup and Sampson, Aaron L. and Lainscsek, Claudia and Muller, Lyle and Lin, Wutu and Doyle, John C. and Cash, Sydney S. and Halgren, Eric and Sejnowski, Terrence J.},
    title = {Interpretation of the Precision Matrix and Its Application in Estimating Sparse Brain Connectivity during Sleep Spindles from Human Electrocorticography Recordings},
    journal = {Neural Computation},
    volume = {29},
    number = {3},
    pages = {603-642},
    year = {2017},
    month = {03},
    issn = {0899-7667},
    doi = {10.1162/NECO_a_00936},
    url = {https://doi.org/10.1162/NECO_a_00936},
}

@inproceedings{Pascanu_difficulty_2013,
author = {Pascanu, Razvan and Mikolov, Tomas and Bengio, Yoshua},
title = {On the difficulty of training recurrent neural networks},
year = {2013},
publisher = {JMLR.org},
booktitle = {Proceedings of the 30th International Conference on International Conference on Machine Learning - Volume 28},
pages = {III–1310–III–1318},
location = {Atlanta, GA, USA},
series = {ICML'13}
}

@ARTICLE{Budd2012wiringCortex,
AUTHOR={Budd, Julian  and Kisvarday, Zoltan F.},
TITLE={Communication and wiring in the cortical connectome},
JOURNAL={Frontiers in Neuroanatomy},
VOLUME={Volume 6 - 2012},
YEAR={2012},
URL={https://www.frontiersin.org/journals/neuroanatomy/articles/10.3389/fnana.2012.00042},
DOI={10.3389/fnana.2012.00042},  
ISSN={1662-5129},
}

@article{miconi_hebb_2017,
	title        = {Biologically plausible learning in recurrent neural networks reproduces neural dynamics observed during cognitive tasks},
	author       = {Miconi, Thomas},
	year         = 2017,
	journal      = {eLife},
	publisher    = {eLife Sciences Publications, Ltd},
	volume       = 6,
	pages        = {e20899},
	doi          = {10.7554/eLife.20899},
	url          = {https://doi.org/10.7554/eLife.20899}
}

@article{fruengel_sparsity_2025,
	title        = {Sparse connectivity enables efficient information processing in cortex-like artificial neural networks},
	author       = {Fruengel, Rieke and Oberlaender, Marcel},
	year         = 2025,
	journal      = {Frontiers in Neural Circuits},
	volume       = 19,
	pages        = 1528309,
	doi          = {10.3389/fncir.2025.1528309},
	issn         = {1662-5110},
	url          = {https://doi.org/10.3389/fncir.2025.1528309}
}

@article{mcallister2026nonrandombrainconnectome,
	title        = {Non-random brain connectome wiring enables robust and efficient neural network function under high sparsity},
	author       = {McAllister, James and Houghton, Conor and Wade, John and O{\textquoteright}Donnell, Cian},
	year         = 2026,
	journal      = {bioRxiv},
	publisher    = {Cold Spring Harbor Laboratory},
	doi          = {10.64898/2026.03.30.715411},
	url          = {https://doi.org/10.64898/2026.03.30.715411}
}

@Article{Reimann2026nonrandomArchitect,
author={Reimann, Michael W.
and Egas Santander, Daniela
and Kanari, Lida
and Barros-Zulaica, Natal{\'i}},
title={Spatial continuity of neurons explains non-random network architecture},
journal={iScience},
year={2026},
month={Jun},
day={19},
publisher={Elsevier},
volume={29},
number={6},
issn={2589-0042},
doi={10.1016/j.isci.2026.116144},
url={https://doi.org/10.1016/j.isci.2026.116144}
}

@Article{Masuda2004smallworldSynch,
author={Masuda, Naoki
and Aihara, Kazuyuki},
title={Global and local synchrony of coupled neurons in small-world networks},
journal={Biological Cybernetics},
year={2004},
month={Apr},
day={01},
volume={90},
number={4},
pages={302-309},
issn={1432-0770},
doi={10.1007/s00422-004-0471-9},
url={https://doi.org/10.1007/s00422-004-0471-9}
}

@article{Chen2017primateconnectome,
    doi = {10.1371/journal.pcbi.1005776},
    author = {Chen, Yuhan AND Wang, Shengjun AND Hilgetag, Claus C. AND Zhou, Changsong},
    journal = {PLOS Computational Biology},
    publisher = {Public Library of Science},
    title = {Features of spatial and functional segregation and integration of the primate connectome revealed by trade-off between wiring cost and efficiency},
    year = {2017},
    month = {09},
    volume = {13},
    url = {https://doi.org/10.1371/journal.pcbi.1005776},
    pages = {1-37},
    number = {9},
}

@article{Chen2006wiringoptimization,
author = {Beth L. Chen  and David H. Hall  and Dmitri B. Chklovskii },
title = {Wiring optimization can relate neuronal structure and function},
journal = {Proceedings of the National Academy of Sciences},
volume = {103},
number = {12},
pages = {4723-4728},
year = {2006},
doi = {10.1073/pnas.0506806103},
URL = {https://www.pnas.org/doi/abs/10.1073/pnas.0506806103},
}

@article{Varshney2011celegansconnectome,
    doi = {10.1371/journal.pcbi.1001066},
    author = {Varshney, Lav R. AND Chen, Beth L. AND Paniagua, Eric AND Hall, David H. AND Chklovskii, Dmitri B.},
    journal = {PLOS Computational Biology},
    publisher = {Public Library of Science},
    title = {Structural Properties of the Caenorhabditis elegans Neuronal Network},
    year = {2011},
    month = {02},
    volume = {7},
    url = {https://doi.org/10.1371/journal.pcbi.1001066},
    pages = {1-21},
    number = {2},
}

@Article{Cherniak1992localoptimizationarbor,
author={Cherniak, Christopher},
title={Local optimization of neuron arbors},
journal={Biological Cybernetics},
year={1992},
month={Apr},
day={01},
volume={66},
number={6},
pages={503-510},
issn={1432-0770},
doi={10.1007/BF00204115},
url={https://doi.org/10.1007/BF00204115}
}

@article{Kaiser2007clustercortical,
title = {Development of multi-cluster cortical networks by time windows for spatial growth},
journal = {Neurocomputing},
volume = {70},
number = {10},
pages = {1829-1832},
year = {2007},
note = {Computational Neuroscience: Trends in Research 2007},
issn = {0925-2312},
doi = {https://doi.org/10.1016/j.neucom.2006.10.060},
url = {https://www.sciencedirect.com/science/article/pii/S0925231206003821},
author = {Marcus Kaiser and Claus C. Hilgetag},
}

@misc{Liao2024selfassemblybiologicallyplausiblelearning,
      title={Self-Assembly of a Biologically Plausible Learning Circuit}, 
      author={Qianli Liao and Liu Ziyin and Yulu Gan and Brian Cheung and Mark Harnett and Tomaso Poggio},
      year={2024},
      eprint={2412.20018},
      archivePrefix={arXiv},
      primaryClass={cs.NE},
      url={https://arxiv.org/abs/2412.20018}, 
}

@Article{Seung2024predictfunctionfromstructure,
author={Seung, H. Sebastian},
title={Predicting visual function by interpreting a neuronal wiring diagram},
journal={Nature},
year={2024},
month={Oct},
day={01},
volume={634},
number={8032},
pages={113-123},
issn={1476-4687},
doi={10.1038/s41586-024-07953-5},
url={https://doi.org/10.1038/s41586-024-07953-5}
}
	
\section*{Supplementary Information}

\renewcommand{\tablename}{Supplementary Table}
\renewcommand{\thetable}{\arabic{table}}
\setcounter{table}{0}

\begin{table*}
	\begin{ruledtabular}
		\begin{tabular}{lccc}
			Model & Task 1 & Task 2 & Task 3 \\ \hline
			\textit{\underline{W*}D*C}   & 0.651 (0.617, 0.684) & $\approx$ 0.50 & $\approx$ 0.50 \\
			\textit{WD*C}    & 0.666 (0.612, 0.719) & 0.952 (0.882, 1.000) & 0.830 (0.769, 0.892) \\
			\textit{WDC}     & 0.629 (0.627, 0.631) & $\approx$ 0.50 & $\approx$ 0.50 \\
			\textit{WD*}     & 0.723 (0.646, 0.799) & $\approx$ 0.50 & $\approx$ 0.50 \\
			\textit{WD}      & 0.631 (0.629, 0.634) & $\approx$ 0.50 & $\approx$ 0.50 \\
			\textit{W\textsubscript{(Simple RNN)}} & 0.631 (0.628, 0.634) & $\approx$ 0.50 & $\approx$ 0.50 \\
			\hline
			\textit{\underline{W*}D*C*}  & 0.722 (0.644, 0.799) & 0.532 (0.481, 0.584) & $\approx$ 0.50 \\
			\textit{WD*C*}   & 0.778 (0.691, 0.865) & $\approx$ 0.50 & $\approx$ 0.50 \\
			\textit{\underline{W*}DC*}   & 0.632 (0.630, 0.634) & 0.557 (0.486, 0.628) & $\approx$ 0.50 \\
			\hline
			\textit{\underline{W!}D*C}   & 0.641 (0.626, 0.656) & $\approx$ 0.50 & $\approx$ 0.50 \\
			\textit{\underline{W!}D*C*}  & 0.687 (0.623, 0.750) & $\approx$ 0.50 & $\approx$ 0.50
		\end{tabular}
	\end{ruledtabular}
	\caption{\label{tab:S1_acc_table_cdf} 
		\textbf{Task accuracy of model variants after empirical cumulative distribution function (ECDF) resampling of $\mathbf{W}_{\mathrm{bio}}$.} Task performance declines substantially after ECDF resampling, indicating that the original biologically derived assignment of weight values contained task-relevant structure beyond the matched marginal distribution alone. Values indicate mean accuracy with 95\% confidence intervals. Although \textit{\underline{W*}} and \textit{\underline{W!}} retain moderate performance in Task~1, they perform near chance in Tasks~2 and~3.
	}
\end{table*}

\begin{table*}
	\begin{ruledtabular}
		\begin{tabular}{lccc}
			Model & Task 1 & Task 2 & Task 3 \\ \hline
			\textit{W*D*C}   & \textbf{0.954} (0.906, 1.000) & \textbf{0.864} (0.824, 0.905) & 0.926 (0.841, 1.000) \\
			\textit{WD*C}    & 0.622 (0.620, 0.623) & 0.816 (0.755, 0.877) & \textbf{1.000} \\
			\textit{WDC}     & 0.620 (0.617, 0.622) & $\approx$ 0.50 & $\approx$ 0.50 \\
			\textit{WD*}     & 0.673 (0.614, 0.732) & $\approx$ 0.50 & $\approx$ 0.50 \\
			\textit{WD}      & 0.620 (0.618, 0.622) & $\approx$ 0.50 & $\approx$ 0.50 \\
			\textit{W\textsubscript{(Simple RNN)}}       & 0.621 (0.619, 0.624) & $\approx$ 0.50 & $\approx$ 0.50 \\
			\hline
			\textit{W*D*C*}  & \textbf{1.000} & \textbf{0.830} (0.765, 0.895) & \textbf{0.976} (0.925, 1.000) \\
			\textit{WD*C*}   & 0.716 (0.637, 0.794) & $\approx$ 0.50 & $\approx$ 0.50 \\
			\textit{W*DC*}   & 0.943 (0.878, 1.000) & 0.799 (0.725, 0.873) & 0.900 (0.805, 0.996) \\
			\hline
			\textit{W!D*C}   & 0.926 (0.861, 0.991) & 0.829 (0.764, 0.894) & 0.975 (0.923, 1.000) \\
			\textit{W!D*C*}  & 1.000 & 0.754 (0.667, 0.841) & 0.950 (0.879, 1.000)
		\end{tabular}
	\end{ruledtabular}
	\caption{\label{tab:S2_acc_table_precision}
		\textbf{Task accuracy when weight initialization is derived from the precision matrix rather than the correlation matrix.} Neuronal constraints were drawn from MICrONS session \underline{6}, scan \underline{6}, field \underline{2}. Values indicate mean accuracy with 95\% confidence intervals across 20 simulation runs over 10 epochs on \textbf{312 nodes}. These results show that multiple function-derived statistics can serve as informative priors for recurrent learning.
	}
\end{table*}

\begin{table*}
	\begin{ruledtabular}
		\begin{tabular}{lccc}
			Model & Task 1 & Task 2 & Task 3 \\ \hline
			\textit{W*D*C}   & \textbf{0.956} (0.905, 1.000) & \textbf{0.849}(0.791, 0.906) & \textbf{1.000} \\
			\textit{WD*C}    & 0.766 (0.687, 0.845) & 0.745 (0.656, 0.834) & 0.949 (0.878, 1.000) \\
			\textit{WDC}     & 0.756 (0.670, 0.842) & $\approx$ 0.50 & $\approx$ 0.50 \\
			\textit{WD*}     & 0.688 (0.629, 0.747) & $\approx$ 0.50 & $\approx$ 0.50 \\
			\textit{WD}      & 0.624 (0.620, 0.627) & $\approx$ 0.50 & $\approx$ 0.50 \\
			\textit{W\textsubscript{(Simple RNN)}}       & 0.623 & $\approx$ 0.50 & $\approx$ 0.50 \\
			\hline
			\textit{W*D*C*}  & 0.975 (0.933, 1.000) & 0.815 (0.745, 0.886) & \textbf{0.950} (0.878, 1.000) \\
			\textit{WD*C*}   & 0.679 (0.614, 0.744) & $\approx$ 0.50 & $\approx$ 0.50 \\
			\textit{W*DC*}  & \textbf{1.000} & \textbf{0.870} (0.840, 0.900) & 0.876 (0.773, 0.979) \\
			\hline
			\textit{W!D*C}   & 0.973 (0.945, 1.000) & 0.848 (0.791, 0.906) & 1.000 \\
			\textit{W!D*C*}  & 1.000 & 0.868 (0.827, 0.910) & 0.962 (0.904, 1.000)
		\end{tabular}
	\end{ruledtabular}
	\caption{\label{tab:S3_acc_table_5_6_8}\textbf{Task accuracy for models built from MICrONS session 5, scan 6, field 8.} Neuronal constraints were drawn from MICrONS \textbf{session \underline{5}, scan \underline{6}, field \underline{8}}. Values are mean accuracy with 95\% confidence intervals across 20 simulation runs over 10 epochs on \textbf{160 nodes}.}
\end{table*}

\begin{table*}
	\begin{ruledtabular}
		\begin{tabular}{lccc}
			Model & Task 1 & Task 2 & Task 3 \\ \hline
			\textit{W*D*C}   & \textbf{0.973} (0.939, 1.000) & 0.869 (0.829, 0.908) & \textbf{1.000} \\
			\textit{WD*C}    & 0.914 (0.847, 0.981) & \textbf{0.885} (0.883, 0.888) & 0.948 (0.876, 1.000) \\
			\textit{WDC}     & 0.900 (0.822, 0.977) & $\approx$ 0.50 & 0.548 $\approx$ 0.50 \\
			\textit{WD*}     & 0.843 (0.763, 0.924) & $\approx$ 0.50 & $\approx$ 0.50 \\
			\textit{WD}      & 0.623 (0.620, 0.626) & $\approx$ 0.50 & $\approx$ 0.50 \\
			\textit{W\textsubscript{(Simple RNN)}}       & 0.624 (0.621, 0.627) & $\approx$ 0.50 & $\approx$ 0.50 \\
			\hline
			\textit{W*D*C*}  & \textbf{0.981} (0.942, 1.000) & 0.826 (0.760, 0.892) & \textbf{0.899} (0.802, 0.996) \\
			\textit{WD*C*}   & 0.898 (0.824, 0.971) & $\approx$ 0.50 & $\approx$ 0.50 \\
			\textit{W*DC*}   & 0.976 (0.942, 1.010) & \textbf{0.863} (0.822, 0.903) & 0.873 (0.767, 0.979) \\
			\hline
			\textit{W!D*C}   & 0.994 (0.980, 1.000) & 0.868 (0.828, 0.908) & 0.950 (0.878, 1.000) \\
			\textit{W!D*C*}  & 1.000 & 0.848 (0.795, 0.901) & 0.950 (0.877, 1.000)
		\end{tabular}
	\end{ruledtabular}
	\caption{\label{tab:S4_acc_table_5_3_4}\textbf{Task accuracy for models built from MICrONS session 5, scan 3, field 4.} Neuronal constraints were drawn from MICrONS \textbf{session \underline{5}, scan \underline{3}, field \underline{4}}. Values are mean accuracy with 95\% confidence intervals across 20 simulation runs over 10 epochs on \textbf{70 nodes}.}
\end{table*}


\subsection*{Statistical decomposition of cortical priors}
\subsubsection{Effect of W for Fixed D, C}

When \textit{D*} was applied, accuracies differed significantly across W variants in all tasks. For Task~1, with D = \textit{D*} and C = \textit{C}, the omnibus test was significant ($H=38.23$, $p=5.0\times10^{-9}$), and pairwise comparisons indicated higher accuracies for \textit{W!} and \textit{W*} than \textit{W} (Holm-adjusted $p=6.16\times10^{-6}$ and $3.81\times10^{-7}$; medians = $0.628$, $1.0$, $1.0$).

A similar pattern emerged for C = \textit{C*} ($H=24.21$, $p=5.55\times10^{-6}$) and persisted in Tasks~2 and~3, where \textit{W!} and \textit{W*} again outperformed \textit{W} under D = \textit{D*} and C = \textit{C*} (Task~2: $H=39.43$, $p=2.74\times10^{-9}$; Task~3: $H=44.11$, $p=2.64\times10^{-10}$).

No significant difference was found between \textit{W!} and \textit{W*}, indicating that the interaction between \textit{W*} and \textit{D*} does not account for the observed performance differences (Supplementary Table~\ref{tab:S5_mw_pairwise_W}).

\subsubsection{Effect of D for Fixed W, C}

With W = \textit{W} and C = \textit{C} fixed, \textit{D*} significantly improved accuracy across tasks comparing to when using \textit{D} (Task~1: $H=11.17$, $p=8.33\times10^{-4}$; Task~2: $H=17.27$, $p=3.25\times10^{-5}$; Task~3: $H=29.76$, $p=4.89\times10^{-8}$). Without D-regularization, differences were non-significant (Supplementary Table~\ref{tab:S6_mw_pairwise_D}).

\subsubsection{Effect of C for Fixed W, D}

Different C formulations produced smaller, task-dependent effects. In Task~1, no pairwise comparisons were significant ($H=1.66$–$5.31$, Holm $p>0.05$).

In Tasks~2 and~3 with D = \textit{D*}, \textit{C} increased accuracy relative to \textit{C*} or the unregularized condition (Task~2: $H=26.44$, $p=1.81\times10^{-6}$; Holm $p<1\times10^{-4}$; Task~3: $H=39.87$, $p=2.2\times10^{-9}$). Smaller but consistent effects were observed for \textit{W!} and \textit{W*} in Task~2 ($H=8.08$–$10.20$, $p<0.005$; Holm $p=0.014$–$0.0059$) (Supplementary Table~\ref{tab:S7_mw_pairwise_C}).

\begin{table*}
	\begin{ruledtabular}
		\centering
		\resizebox{\textwidth}{!}{%
			\begin{tabular}{lccccc}
				Task & Fixed (D, C) & Comparison & $U$ & Holm $p$ & Median direction\\ \hline
				
				\multirow{6}{*}{1}
				& \multirow{3}{*}{(\textit{D*}, \textit{C})}
				& \textit{W} -- \textit{W!}    & $26.0$  & \bm{$6.16\times10^{-6}$} & $\textit{W} < \textit{W!}$ \\
				&  & \textit{W} -- \textit{W*}  & $3.0$   & \bm{$3.81\times10^{-7}$} & $\textit{W} < \textit{W*}$ \\
				&  & \textit{W!} -- \textit{W*} & $226.0$ & $0.761$             & $\textit{W!} = \textit{W*}$ \\
				\cmidrule(lr){2-6}
				& \multirow{3}{*}{(\textit{D*}, \textit{C*})}
				& \textit{W} -- \textit{W!}    & $83.0$  & \bm{$8.70\times10^{-4}$} & $\textit{W} < \textit{W!}$ \\
				&  & \textit{W} -- \textit{W*}  & $73.5$  & \bm{$2.52\times10^{-4}$} & $\textit{W} < \textit{W*}$ \\
				&  & \textit{W!} -- \textit{W*} & $189.0$ & $0.761$             & $\textit{W!} = \textit{W*}$ \\ \hline
				
				\multirow{6}{*}{2}
				& \multirow{3}{*}{(\textit{D*}, \textit{C})}
				& \textit{W} -- \textit{W!}    & $49.5$  & \bm{$1.97\times10^{-4}$} & $\textit{W} < \textit{W!}$ \\
				&  & \textit{W} -- \textit{W*}  & $51.5$  & \bm{$1.97\times10^{-4}$} & $\textit{W} < \textit{W*}$ \\
				&  & \textit{W!} -- \textit{W*} & $235.0$ & $0.701$             & $\textit{W!} > \textit{W*}$ \\
				\cmidrule(lr){2-6}
				& \multirow{3}{*}{(\textit{D*}, \textit{C*})}
				& \textit{W} -- \textit{W!}    & $0.0$   & \bm{$4.02\times10^{-7}$} & $\textit{W} < \textit{W!}$ \\
				&  & \textit{W} -- \textit{W*}  & $0.0$   & \bm{$4.02\times10^{-7}$} & $\textit{W} < \textit{W*}$ \\
				&  & \textit{W!} -- \textit{W*} & $213.5$ & $0.725$             & $\textit{W!} > \textit{W*}$ \\ \hline
				
				\multirow{3}{*}{3}
				& \multirow{3}{*}{(\textit{D*}, \textit{C*})}
				& \textit{W} -- \textit{W!}    & $14.0$  & \bm{$2.75\times10^{-7}$} & $\textit{W} < \textit{W!}$ \\
				&  & \textit{W} -- \textit{W*}  & $14.0$  & \bm{$2.75\times10^{-7}$} & $\textit{W} < \textit{W*}$ \\
				&  & \textit{W!} -- \textit{W*} & $200.0$ & $1.0$               & $\textit{W!} = \textit{W*}$ \\
				
			\end{tabular}
		}
	\end{ruledtabular}
	\caption{\label{tab:S5_mw_pairwise_W}
		\textbf{Pairwise post-hoc Mann--Whitney tests across \textit{W} variants (Holm-adjusted), grouped by fixed (\textit{D}, \textit{C}) settings.}
		In Task~3, the Kruskal--Wallis test was not significant for the (\textit{D*}, \textit{C}) setting ($p = 0.926$); accordingly, no post-hoc tests were performed for that condition. Across tasks, biologically informed weight-initialized variants (\textit{W!}, \textit{W*}) consistently outperform the baseline setting (\textit{W}). The \textit{Median direction} column indicates which variant had the higher median performance, and the Holm-adjusted $p$-value indicates whether that difference was statistically significant.
	}
\end{table*}

\begin{table*}
	\begin{ruledtabular}
		\centering
		\resizebox{\textwidth}{!}{
			\begin{tabular}{lccccc}
				Task & Fixed (W, C) & Comparison & $U$ & Holm $p$ & Median direction\\ \hline
				
				\multirow{5}{*}{1}
				& \multirow{1}{*}{(\textit{W}, \textit{C})}
				& \textit{D} -- \textit{D*}        & $76.5$  & \bm{$4.37\times10^{-3}$} & $\textit{D} < \textit{D*}$ \\
				\cmidrule(lr){2-6}
				& \multirow{3}{*}{(\textit{W})}
				& \textit{D} -- \textit{D*}        & $138.0$ & $0.285$             & $\textit{D} < \textit{D*}$ \\
				&  & \textit{D} -- \textit{\cancel{D}}              & $262.0$ & $0.285$             & $\textit{D} > \textit{\cancel{D}}$ \\
				&  & \textit{D*} -- \textit{\cancel{D}}             & $303.0$ & \bm{$2.15\times10^{-2}$} & $\textit{D*} > \textit{\cancel{D}}$ \\
				\cmidrule(lr){2-6}
				& \multirow{1}{*}{(\textit{W*}, \textit{C*})}
				& \textit{D} -- \textit{D*}        & $180.5$ & $0.324$             & $\textit{D} = \textit{D*}$ \\ \hline
				
				\multirow{2}{*}{2}
				& \multirow{1}{*}{(\textit{W}, \textit{C})}
				& \textit{D} -- \textit{D*}        & $46.5$  & \bm{$6.89\times10^{-5}$} & $\textit{D} < \textit{D*}$ \\
				\cmidrule(lr){2-6}
				& \multirow{1}{*}{(\textit{W*}, \textit{C*})}
				& \textit{D} -- \textit{D*}        & $191.0$ & $0.818$             & $\textit{D} < \textit{D*}$ \\ \hline
				
				\multirow{2}{*}{3}
				& \multirow{1}{*}{(\textit{W}, \textit{C})}
				& \textit{D} -- \textit{D*}        & $6.5$   & \bm{$1.06\times10^{-7}$} & $\textit{D} < \textit{D*}$ \\
				\cmidrule(lr){2-6}
				& \multirow{1}{*}{(\textit{W*}, \textit{C*})}
				& \textit{D} -- \textit{D*}        & $189.0$ & $0.621$             & $\textit{D} = \textit{D*}$ \\
				
			\end{tabular}
		}
	\end{ruledtabular}
	\caption{\label{tab:S6_mw_pairwise_D}
		\textbf{Pairwise post-hoc Mann--Whitney tests across \textit{D} variants (Holm-adjusted), grouped by fixed (\textit{W}, \textit{C}) settings.}
		For the baseline \textit{W} setting, the Kruskal--Wallis test was not significant in Task~2 ($p = 0.10$) or Task~3 ($p = 0.349$); accordingly, no post-hoc tests were performed for those conditions. Overall, \textit{D*} variants tend to show higher accuracies than \textit{D} variants under baseline settings, especially for $(\textit{W}, \textit{C})$, although this pattern does not hold for $(\textit{W*}, \textit{C*})$. ``\textit{\cancel{D}}'' denotes that spatial embedding was not applied for that setting.
	}
\end{table*}

\begin{table*}
	\begin{ruledtabular}
		\centering
		\resizebox{\textwidth}{!}{
			\begin{tabular}{lccccc}
				Task & Fixed (W, D) & Comparison & $U$ & Holm $p$ & Median direction\\ \hline
				
				\multirow{3}{*}{1}
				& (\textit{W}, \textit{D}) 
				& \textit{C} -- \textit{\cancel{C}}           & $114.5$ & $\bm{6.42\times10^{-2}}$ & $\textit{C} < \text{\textit{\cancel{C}}}$ \\
				\cmidrule(lr){2-6}
				& (\textit{W!}, \textit{D*}) 
				& \textit{C} -- \textit{C*}   & $181.0$ & $0.421$              & $\textit{C} = \textit{C*}$ \\
				\cmidrule(lr){2-6}
				& (\textit{W*}, \textit{D*}) 
				& \textit{C} -- \textit{C*}   & $140.5$ & $\bm{6.42\times10^{-2}}$ & $\textit{C} = \textit{C*}$ \\ \hline
				
				\multirow{6}{*}{2}
				& (\textit{W}, \textit{D}) 
				& \textit{C} -- \textit{\cancel{C}}           & $268.0$ & $0.135$              & $\textit{C} > \text{\textit{\cancel{C}}}$ \\
				\cmidrule(lr){2-6}
				& \multirow{3}{*}{(\textit{W}, \textit{D*})}
				& \textit{C} -- \textit{C*}   & $366.0$ & \bm{$4.52\times10^{-5}$} & $\textit{C} > \textit{C*}$ \\
				& & \textit{C} -- \textit{\cancel{C}}         & $358.5$ & \bm{$9.53\times10^{-5}$} & $\textit{C} > \text{\textit{\cancel{C}}}$ \\
				& & \textit{C*} -- \textit{\cancel{C}}        & $164.0$ & $0.336$              & $\textit{C*} < \text{\textit{\cancel{C}}}$ \\
				\cmidrule(lr){2-6}
				& (\textit{W!}, \textit{D*}) 
				& \textit{C} -- \textit{C*}   & $305.0$ & \bm{$1.40\times10^{-2}$} & $\textit{C} > \textit{C*}$ \\
				\cmidrule(lr){2-6}
				& (\textit{W*}, \textit{D*}) 
				& \textit{C} -- \textit{C*}   & $318.0$ & \bm{$5.90\times10^{-3}$} & $\textit{C} > \textit{C*}$ \\ \hline
				
				\multirow{6}{*}{3}
				& (\textit{W}, \textit{D}) 
				& \textit{C} -- \textit{\cancel{C}}           & $187.0$ & $1.0$                & $\textit{C} = \text{\textit{\cancel{C}}}$ \\
				\cmidrule(lr){2-6}
				& \multirow{3}{*}{(\textit{W}, \textit{D*})}
				& \textit{C} -- \textit{C*}   & $393.0$ & \bm{$3.15\times10^{-7}$} & $\textit{C} > \textit{C*}$ \\
				& & \textit{C} -- \textit{\cancel{C}}         & $390.0$ & \bm{$3.38\times10^{-7}$} & $\textit{C} > \text{\textit{\cancel{C}}}$ \\
				& & \textit{C*} -- \textit{\cancel{C}}        & $152.5$ & $0.654$              & $\textit{C*} = \text{\textit{\cancel{C}}}$ \\
				\cmidrule(lr){2-6}
				& (\textit{W!}, \textit{D*}) 
				& \textit{C} -- \textit{C*}   & $192.0$ & $1.0$                & $\textit{C} = \textit{C*}$ \\
				\cmidrule(lr){2-6}
				& (\textit{W*}, \textit{D*}) 
				& \textit{C} -- \textit{C*}   & $190.0$ & $1.0$                & $\textit{C} = \textit{C*}$ \\
				
			\end{tabular}
		}
	\end{ruledtabular}
	\caption{\label{tab:S7_mw_pairwise_C}
		\textbf{Pairwise post-hoc Mann--Whitney tests across \textit{C} variants (Holm-adjusted), grouped by fixed (\textit{W}, \textit{D}) settings.}
		In Task~1, the Kruskal--Wallis test was not significant for the (\textit{W}, \textit{D*}) setting ($p = 0.436$); accordingly, no post-hoc tests were performed for that condition. The effect of \textit{C} is modest overall but becomes more pronounced when combined with \textit{D*} in Tasks~2 and~3. ``\textit{\cancel{C}}'' denotes that communicability calculations were not carried out for that setting.
	}
\end{table*}

\begin{table*}
	\begin{ruledtabular}
		\begin{tabular}{lccc}
			\multicolumn{4}{c}{\textbf{One-choice Inference}} \\
			\midrule[0.6pt]
			Problem & Stimulus A & Stimulus B & Correct Decision \\
			\midrule[0.3pt]
			1  & (Right, up)   & Left, right & Right \\
			2  & (Right, up)   & Right, down & Right \\
			3  & (Right, up)   & Up, down    & Up \\
			4  & (Right, up)   & Up, left    & Up \\
			5  & (Right, down) & Left, right & Right \\
			6  & (Right, down) & Up, right   & Right \\
			7  & (Right, down) & Up, down    & Down \\
			8  & (Right, down) & Left, down  & Down \\
			9  & (Left, up)    & Left, right & Left \\
			10 & (Left, up)    & Left, down  & Left \\
			11 & (Left, up)    & Up, down    & Up \\
			12 & (Left, up)    & Up, right   & Up \\
			13 & (Left, down)  & Left, right & Left \\
			14 & (Left, down)  & Up, left    & Left \\
			15 & (Left, down)  & Up, down    & Down \\
			16 & (Left, down)  & Right, down & Down \\[6pt]
			
			\multicolumn{4}{c}{\textbf{Perceptual Decision-making}} \\
			\midrule[0.6pt]
			Problem & Coherence (\%) & Stimulus Direction & Correct Decision \\
			\midrule[0.3pt]
			1  & 0    & Right & Right \\
			2  & 6.4  & Right & Right \\
			3  & 12.8 & Right & Right \\
			4  & 25.6 & Right & Right \\
			5  & 51.2 & Right & Right \\
			6  & 0    & Left  & Left \\
			7  & 6.4  & Left  & Left \\
			8  & 12.8 & Left  & Left \\
			9  & 25.6 & Left  & Left \\
			10 & 51.2 & Left  & Left \\[6pt]
			
			\multicolumn{4}{c}{\textbf{Go/NoGo}} \\
			\midrule[0.6pt]
			Problem & Description & Stimulus Encoding & Correct Decision \\
			\midrule[0.3pt]
			1 & NoGo & [0, 1, 0] & [1, 0] (NoGo) \\
			2 & Go   & [0, 0, 1] & [0, 1] (Go)   \\
			
		\end{tabular}
	\end{ruledtabular}
	\caption{\label{tab:S8_all_tasks}\textbf{All task instances used to train and evaluate the networks.}}
\end{table*}

\begin{table*}
	\begin{ruledtabular}
		\begin{tabular}{lcccc}
			Session\_Scan\_Field & Neurons & Grid & Intervals / Total Time \\ \hline
			5\_3\_4 & 70  & (7, 5, 2)   & Clip, Monet2, Trippy / 1.76 h \\
			5\_6\_8 & 160 & (10, 4, 4)  & Clip, Monet2, Trippy / 1.76 h \\
			6\_6\_2 & 312 & (12, 13, 2) & Clip, Monet2, Trippy / 1.76 h \\
		\end{tabular}
	\end{ruledtabular}
	\caption{\label{tab:S9_session_scan_fields}\textbf{MICrONS fields used for spatial embedding and functional weight initialization.} Each row corresponds to one session\_scan\_field combination. \textbf{Neurons} gives the number of neurons used as network nodes, and \textbf{Grid} gives the spatial layout used to match network size in random spatial models. Functional metrics were computed from calcium imaging response to \textit{Clip}, \textit{Monet2}, and \textit{Trippy} stimuli \cite{bae2025functionalconnectomicsspanning}, totaling 1.76 h of data.}
\end{table*}

\end{document}